\documentclass[a4paper,fleqn]{cas-dc}
\usepackage[authoryear,longnamesfirst]{natbib}
\usepackage{algorithm} 
\usepackage{algpseudocode} 
\usepackage{makecell}
\usepackage{arydshln}
\usepackage{array}
\usepackage{booktabs}
\usepackage {subcaption}
\usepackage{colortbl}
\usepackage{placeins}
\usepackage[normalem]{ulem}
\usepackage{float}

\def\tsc#1{\csdef{#1}{\textsc{\lowercase{#1}}\xspace}}
\tsc{WGM}
\tsc{QE}
\tsc{EP}
\tsc{PMS}
\tsc{BEC}
\tsc{DE}
\begin{document}
\let\WriteBookmarks\relax
\def\floatpagepagefraction{1}
\def\textpagefraction{.001}
\definecolor{mygray}{gray}{.8}

\title [mode = title]{An Efficient Multi-Indicator and Many-Objective Optimization Algorithm based on Two-Archive}

\author[1]{Ziming Wang}[style=chinese]

\ead {12132363@mail.sustech.edu.cn}

\affiliation[1]{organization={Guangdong Key Laboratory of Brain-Inspired Intelligent Computation},
    addressline={Department of Computer Science and Engineering}, 
    city={shenzhen},
    postcode={518055}, 
    country={China}}

\author[1]{Xin Yao}[style=chinese]

\cormark[1]

% Email id of the second author
\ead{xiny@sustech.edu.cn}

\cortext[cor1]{Corresponding author}
% \cortext{E-mail address}

\begin{abstract}
Indicator-based algorithms are gaining prominence as traditional multi-objective optimization algorithms based on domination and decomposition struggle to solve many-objective optimization problems. However, previous indicator-based multi-objective optimization algorithms suffer from the following flaws: 1) The environment selection process takes a long time; 2) Additional parameters are usually necessary. As a result, this paper proposed an multi-indicator and multi-objective optimization algorithm based on two-archive (SRA3) that can efficiently select good individuals in environment selection based on indicators performance and uses an adaptive parameter strategy for parental selection without setting additional parameters. Then we normalized the algorithm and compared its performance before and after normalization, finding that normalization improved the algorithm's performance significantly. We also analyzed how normalizing affected the indicator-based algorithm and observed that the normalized $I_{\epsilon+}$ indicator is better at finding extreme solutions and can reduce the influence of each objective's different extent of contribution to the indicator due to its different scope. However, it also has a preference for extreme solutions, which causes the solution set to converge to the extremes. As a result, we give some suggestions for normalization. Then, on the DTLZ and WFG problems, we conducted experiments on 39 problems with 5, 10, and 15 objectives, and the results show that SRA3 has good convergence and diversity while maintaining high efficiency. Finally, we conducted experiments on the DTLZ and WFG problems with 20 and 25 objectives and found that the algorithm proposed in this paper is more competitive than other algorithms as the number of objectives increases.
\end{abstract}

% Keywords

\maketitle

\section{Introduction}

Multi-objective optimization problems (MOPs) with more than three objectives are called many-objective optimization problems (MaOPs) \cite{fleming2005many,Praditwong2007how}. There are many MaOPs in real life \cite{2004Applications,2011Software,2012Many}, but traditional multi-objective evolutionary algorithms (MOEAs) based on domination and decomposition have difficulty in solving MaOPs \cite{khare2003performance}. In contrast, indicator-based multi-objective optimization algorithms (IB-MOEAs) use indicator values to direct the search process when solving MaOPs. This method of evaluating indicators directly to search well-performing solutions is intuitive and gaining popularity. There are various indicators used in IB-MOEAs  \cite{li2019quality}, such as hypervolume (HV) \cite{zitzler1998multiobjective}, generational distance (GD) \cite{van1998evolutionary}, inverted generational distance (IGD) \cite{coello2004study}, epsilon indicator ($\varepsilon$-indicator) \cite{zitzler2003performance}, shift-based density estimation (SDE) \cite{li2013shift}, R2 \cite{hansen1994evaluating}, $\Delta_p$ \cite{schutze2012using} and so on.

However, algorithms based on a single indicator may be biased in the search process \cite{falcon2020indicator,2011Boosting}. For example, indicator-based evolutionary algorithm (IBEA) \cite{zitzler2004indicator} that uses ${I_{\varepsilon+}}$ indicator has difficulty maintaining diversity when solving MaOPs. And previous algorithms based on multi-indicators or combining indicators with other approaches may be more time-consuming and usually require setting of additional parameters. For example, stochastic ranking algorithm (SRA) \cite{li2016stochastic} uses a stochastic ranking technique in environment selection, which requires a large number of exchanges and takes a long time, as well as setting the parameter $P_c$ for selecting indicators, while an improved two-archive algorithm (Two\_Arch2) \cite{wang2014two_arch2} requires a higher time cost in the update of DA archive, as well as setting the number of CA archive and the parameter $p$ for calculating the distance. Because of this, we propose a multi-indicator and multi-objective optimization algorithm based on two-archive (SRA3) and demonstrate through extensive experiments that this algorithm has better performance.

The main contributions of this paper are summarized as follows.
\begin{enumerate}[(1)]
\item An efficient two-archive based environment selection strategy is adopted.
\item An adaptive strategy for parental selection is given without using any additional parameters.
\item The impact of normalizing on several current indicator-based algorithms is investigated, and some suggestions for normalization guidelines are given.
\item Scalability analysis was performed for several algorithms and it was found that SRA3 gradually has advantages as the number of objectives increases.
\end{enumerate}

The rest of this paper is organized as follows. Section 2 presents the details of SRA3. Section 3 describes the experimental setup, the experimental study, and the analysis. Section 4 summarizes the paper and indicates the direction of future research.

\section{An efficient multi-indicator and many-objective optimization algorithm based on two-archive}

% This section will include a description of the overall architecture of the algorithm, how the archives are updated, and an analysis of the algorithm efficiency.

\subsection{Overview}
In the past, MOEAs based on two-archive using CA and DA archives to maintain the convergence and diversity of solution set respectively have been proved to have desired results \cite{wang2014two_arch2,li2018two,2006A}. Therefore, based on the two-archive framework, this paper proposes a multi-indicator multi-objective optimization algorithm (SRA3). First, $N$ solutions are randomly generated as the initial archives of CA and DA. In each generation, solutions are adaptively selected from CA and DA as parents to generate offspring based on the proportion of non-dominated solutions in the CA and DA archives. Then the fitness and indicator values of the offspring are calculated, and the CA and DA archives are updated based on the indicator performance of the offspring. When the iteration is completed, the non-dominated solutions in the CA archive are returned as the output. The framework of SRA3 is described in Algorithm 1.

\renewcommand{\algorithmicrequire}{\textbf{Input:}}  % Use Input in the format of Algorithm 1  
\renewcommand{\algorithmicensure}{\textbf{Output:}} % Use Output in the format of Algorithm 1
\begin{algorithm}
	\caption{Main Loop of SRA3} 
	\begin{algorithmic}[1]
	\Require  an MaOP, archive size N
    \Ensure  an approximate set $P_{out}$
		\State Random generation of initial archives CA and DA
		\State Evaluate all the solutions in CA and DA
		\State $t\leftarrow0$
		\While {$t\textless maxGen$}
		    \State $Q_t\leftarrow$\textit{generateOffspring}$(CA,DA)$
		    \State Evaluate all the solutions in $Q_t$
		    \State $CA\leftarrow updateCA(CA\cup Q_t)$
		    \State $DA\leftarrow updateDA(DA\cup Q_t)$
		    \State $t\leftarrow{t+1}$
		\EndWhile
		\State $P_{out}\leftarrow$ non-dominated solutions in CA
		\State \Return $P_{out}$
	\end{algorithmic} 
\end{algorithm}

\subsection{Adaptive parental selection and generation of offspring}
One of the SRA3's features is the ability to make adaptive parental selection. Its selection is based on the proportion of non-dominated solutions in the CA and DA archives. Specifically, first, the proportions $p_c$ and $p_d$ of non-dominated solutions in CA and DA archives are calculated separately, and if $p_c$ is greater than $p_d$, an individual is randomly selected from CA as parent 1, otherwise an individual is randomly selected from DA. Then, CA and DA are combined into a set S, and then the proportions $\rho_c$ and $\rho_d$ of the non-dominated solutions of CA and DA in S are evaluated separately. The proportion of $\rho_c$ in the sum of $\rho_c$ and $\rho_d$ determines whether parent 2 chooses from CA or DA, and the larger the proportion of $\rho_c$, the higher the probability of selecting an individual from CA as parent 2. After parental selection, simulated binary crossover (SBX) \cite{deb1995simulated} and polynomial mutation \cite{1996A} are used to generate offspring. The pseudo-code for adaptive parent selection and offspring generation are shown in Algorithm 2.

\renewcommand{\algorithmicrequire}{\textbf{Input:}}  % Use Input in the format of Algorithm 2  
\renewcommand{\algorithmicensure}{\textbf{Output:}} % Use Output in the format of Algorithm 2
\begin{algorithm}
	\caption{\textit{generateOffspring(CA,DA)} procedure} 
	\begin{algorithmic}[1]
	\Require  CA archive, DA archive
    \Ensure  offspring population Q
		\State $i\leftarrow0,Q\leftarrow\emptyset$
		\State $p_c\leftarrow$proportion of non-dominated solutions in CA
		\State $p_d\leftarrow$proportion of non-dominated solutions in DA
		\State $S\leftarrow{CA\cup{DA}}$
		\State $\rho_c\leftarrow$proportion of non-dominated solutions of CA in S
		\State $\rho_d\leftarrow$proportion of non-dominated solutions of DA in S
		\While{$i\textless{len(CA)}$}
		    \If{$p_c\textgreater{p_d}$}
		        \State $p1\leftarrow{randSelect(CA)}$
		    \Else
		        \State $p1\leftarrow{randSelect(DA)}$
		    \EndIf
		    \If{$rand()\textless{\rho_c/(\rho_c+\rho_d)}$}
		        \State $p2\leftarrow{randSelect(CA)}$
		    \Else
		        \State $p2\leftarrow{randSelect(DA)}$
		    \EndIf
		    \State create an offspring $q_i$ with $p1$ and $p2$
		    \State $Q\leftarrow{Q\cup{q_i}}$
		    \State $i\leftarrow{i+1}$
		\EndWhile
		\State \Return Q
	\end{algorithmic} 
\end{algorithm}

\subsection{Archives update}

% The quality indicators used and the updates of the CA and DA archives are shown below.

\subsubsection{Update of CA archive}
The $\epsilon$-indicator has been found to have a better performance in maintaining convergence, so SRA3 choose the quality indicator $I_{\epsilon+}$ as the selection principle for CA with reference to IBEA and SRA \cite{zitzler2004indicator,li2016stochastic}. The indicator $I_{\epsilon+}$ measures the minimal distance that a solution should be translated in each dimension to weakly Pareto dominate the other solution \cite{falcon2020indicator}, which is shown by equation (1). Then we refer to the IBEA's approach to calculate the individual fitness as shown in equation (2).

\begin{equation}
    I_{\epsilon+}(x,y)=min_\epsilon(f_i(x)-\epsilon\leq{f_i(y)},i\in\{1,2,\dots,m\})
\end{equation}

\begin{equation}
    I_1(x)=\sum_{y\in{P},y\neq{x}}-e^{-I_{\epsilon+}(y,x)/k}
\end{equation}
where $m$ is the number of objectives, $k$ is a scaling factor that must be greater than 0 and depends on $I_{\epsilon+}$ and the underlying problem, and $P$ is the population that includes $x$ and $y$.

In the step of updating the CA archive of SRA3, the input is the union set $H_1$ of the original CA archive and the offspring population $Q_t$, then the quality indicator value $I_1(x)$ of all individuals in $H_1$ is calculated, and the $N$ individuals with the best performance of the $I_1(x)$ value are selected as the new CA archive, where $N$ is the size of the CA archive. The pseudo-code for CA archive updating is shown in Algorithm 3.

\renewcommand{\algorithmicrequire}{\textbf{Input:}}  % Use Input in the format of Algorithm 2  
\renewcommand{\algorithmicensure}{\textbf{Output:}} % Use Output in the format of Algorithm 2
\begin{algorithm}
	\caption{\textit{updateCA($H_1$)} procedure} 
	\begin{algorithmic}[1]
	\Require  $H_1$ (the union set of CA and $Q_t$)
    \Ensure  the updated CA
		\State compute the indicator values $I_1(h_i)$ for all $h_i\in{H_1}$
		\State the $N$ individuals with the best performance of quality indicator value $I_1(h_i)$ in $H_1$ are selected as the new CA
		\State \Return CA
	\end{algorithmic} 
\end{algorithm}

\subsubsection{Update of DA archive}

$I_{SDE}$ indicator prefers diversity compared to convergence, so it was used as a selection principle to update the DA archive and the $I_{SDE}$ indicator is shown in equations 3, 4, and 5.

\begin{equation}
    I_{SDE}(x,y)=\sqrt{\sum_{1\leq{i}\leq{m}}dist(f_i(x),f_i(y))^2}
\end{equation}

\begin{equation}
    I_2(x)=\frac{\sum_{y\in{P},y\neq{x}}I_{SDE}(x,y)}{2N-1}
\end{equation}
where

\begin{equation}
    dist(f_i(x),f_i(y))=
    \begin{cases}
    f_i(y)-f_i(x)& \text{if $f_i(x)\textless{f_i(y)}$}\\
    0& \text{otherwise}
    \end{cases}
\end{equation}
where $m$ is the number of objectives, $N$ is the size of archive and $P$ is the population that includes $x$ and $y$.

In the step of updating the DA archive of SRA3, the input is the union set $H_2$ of the original DA archive and the offspring population $Q_t$, then the quality indicator value $I_2(x)$ of all individuals in $H_2$ is calculated, and the $N$ individuals with the best performance of the $I_2(x)$ value are selected as the new DA archive, where $N$ is the size of the DA archive. The pseudo-code for DA archive updating is shown in Algorithm 4.

\renewcommand{\algorithmicrequire}{\textbf{Input:}}  % Use Input in the format of Algorithm 2  
\renewcommand{\algorithmicensure}{\textbf{Output:}} % Use Output in the format of Algorithm 2
\begin{algorithm}
	\caption{\textit{updateDA($H_2$)} procedure} 
	\begin{algorithmic}[1]
	\Require  $H_2$ (the union set of DA and $Q_t$)
    \Ensure  the updated DA
		\State compute the indicator values $I_2(h_i)$ for all $h_i\in{H_2}$
		\State the $N$ individuals with the best performance of quality index value $I_2(h_i)$ in $H_2$ are selected as the new DA
		\State \Return DA
	\end{algorithmic} 
\end{algorithm}

\subsection{Algorithm efficiency analysis}

We found that SRA takes a long time in practice due to its stochastic ranking technique in environment selection, which requires frequent swapping and the computational complexity is $O(N^2)$. In contrast, SRA3 directly selects the $N$ individuals with the best quality indicator performance in environment selection, and its computational complexity is $O(NlogN)$.

\section{Experimental study and discussion}

In Section 3, we conducted experiments to verify the performance of SRA3. First we performed the relevant experimental setup, then we compared SRA3 with some state-of-the-art algorithms on each problem, analyzed the effect of normalization on the indicator-based algorithms, and explored the performance trend of some algorithms as the number of objectives increases.

\subsection{Experimental setup}

In Section 3.1, we undertook the necessary pre-experiment preparations, such as selecting the test problems, choosing quality indicators for evaluating the performance, determining the algorithms used for comparison, and specifying setting of parameters for each algorithm and indicator.

\subsubsection{Test problems}

The SRA states that experiments should be conducted on both the DTLZ \cite{2002Scalable} and WFG \cite{1705400} problems because they have very different characteristics \cite{li2016stochastic}. Therefore, referring to the SRA, DTLZ1-DTLZ4 and WFG1-WFG9 are selected for empirical studies and the number of objectives $m$ are 5, 10, 15, 20, and 25. $k$ is set to 5 for DTLZ1, 10 for DTLZ2-DTLZ4, and the number of variables is $m+k-1$ for DTLZ1-DTLZ4 problems. For the WFG1-WFG9 problems, $k$ is set to $m-1$, $l$ is set to 10, and the number of variables is $k+l$.

\subsubsection{Performance metrics}

In the experiments, we employed two widely used quality indicators as evaluation metrics, namely HV \cite{zitzler1998multiobjective} and IGD \cite{coello2004study} indicators.

The HV value of the set $P$ is defined as the volume of the objective space dominated by solutions in $P$ and bounded by $z^*$ \cite{1998Multiobjective}, which is defined as:

\begin{equation}
    HV(P,z^*)=\lambda(\mathop{\bigcup}\limits_{p\in{P}}\{x|p\prec{x}\prec{z^*}\})
\end{equation}
where $\lambda$ is the Lebesgue measure and $z^*$ is the reference point in the objective space. Before calculating the value of HV, the set $P$ is normalized using $1.1\times{z^{nadir}}$, and then the reference point $z^*$ is set to $(1.0, 1.0,\ldots,1.0)^T$.

The IGD value of the set $P$ describes the distance from the reference set $R$ to the set $P$ \cite{2018Reference}, which is defined as:

\begin{equation}
    IGD(P,R)=\frac{1}{|R|}\sum_{i=1}^{|R|}\min_{p\in{P}}dist(r_i,p)
\end{equation}
where $dist(r_i,p)$ denotes the Euclidean distance between $r_i$ and $p$, and $P$ is a subset of the true Pareto front. Since the true Pareto front is known for the DTLZ1-DTLZ4 and WFG1-WFG9 problems, we sampled 500$\,$000 points uniformly on the Pareto surface as the reference set $R$ \cite{8477730}. In addition, we normalized the set $P$ and the reference set $R$ before calculating the IGD indicator.

\subsubsection{State-of-the-Art Algorithms}

In order to verify the performance of SRA3, we selected many state-of-the-art MOEAs for comparison, such as SRA, Two$\_$Arch2, IBEA, NSGA-III, and MOEA/D. Among them, SRA \cite{li2016stochastic} is the well-performing multi-indicator-based algorithm, Two$\_$Arch2 \cite{wang2014two_arch2} uses the same two-archive framework as this paper and also uses $I_{\epsilon+}$ indicator in the CA archive, IBEA \cite{zitzler2004indicator} is the most classical single-indicator-based algorithm, NSGA-III \cite{deb2013evolutionary} is a widely used dominance-based algorithm for solving many objects, and MOEA/D \cite{zhang2007moea} is the classical decomposition-based algorithm.

\subsubsection{Parameter settings}

The general and algorithm-specific parameter settings are shown below.

\begin{enumerate}[(1)]
\item \textit{Population Size}: The number of the population in each algorithm is shown in Table 1. In addition, the CA archive size for Two$\_$Arch2 is set to 100, as recommended by \cite{wang2014two_arch2}.

\begin{table}[width=.9\linewidth,cols=4,pos=h]
\caption{Setting the population size for all algorithms.}\label{tbl1}
\begin{tabular*}{\tblwidth}{@{} CCCC@{} }
\toprule
$m$ & No. of vector & NSGA-III & Other algorithms \\
\midrule
5 & 210 & 212 & 210 \\
10 & 275 & 276 & 275 \\
15 & 135 & 136 & 135 \\
20 & - & - & 135 \\
25 & - & - & 135 \\
\bottomrule
\end{tabular*}
\end{table}

\item \textit{Reproduction Operators}: The simulated binary crossover operator (SBX) and polynomial mutation are used for reproducing offspring solutions \cite{deb2000efficient}. The probability of crossover is set to 1.0, and the probability of mutation is set to $\frac{1}{n}$, where $n$ is the number of variables, and the crossover distribution index $\eta_c$ and the mutation distribution index $\eta_m$ are both set to 20.0 \cite{2002A,8027123}.

\item \textit{Number of iterations and runs}: On each problem, all algorithms are allowed for a maximum of 90$\,$000 fitness evaluations and repeated 20 times independently.

\item \textit{Parameter settings for each algorithm}: For SRA3, the parameter $k$ in indicator $I_{\epsilon+}$ is set to 0.025. For SRA, the parameter $k$ in indicator $I_{\epsilon+}$ is set to 0.05 and the parameter $p_c$ is set to a random number in the range of [0.4,0.6] \cite{li2016stochastic}. For Two$\_$Arch2, the parameter $k$ in indicator $I_{\epsilon+}$ is set to 0.05, the size of the CA archive, as mentioned above, is set to 100 and the parameter $p$ required to calculate the distance is set to $\frac{1}{m}$, where $m$ is the number of objectives \cite{wang2014two_arch2}. For IBEA, the parameter $k$ in indicator $I_{\epsilon+}$ is set to 0.05 \cite{zitzler2004indicator}. For MOEA/D, the number of neighbors is set to $\lceil{N/10}\rceil$, where $N$ is the size of the population and the maximum replacement number is set to 2 \cite{zhang2007moea}. Each algorithm set the parameters as suggested in the original papers.

\item \textit{Statistical Test}: In order to test the difference between the algorithms, the Wilcoxon's rank sum test with a 0.05 significance level is used for analysis \cite{1944Individual}.

\end{enumerate}

\subsection{Experimental study of SRA3 on DTLZ and WFG problems}

On the DTLZ1-4 and WFG1-9 problems with 5, 10, and 15 objectives, we compared SRA3 with the SRA algorithm, which uses multiple indicators as well, and the Two$\_$Arch2 algorithm, which also uses a two-archive framework, in order to verify the performance of SRA3 on each problem. The HV and IGD results are shown in Tables 2 to 5.

\begin{table}[width=.9\linewidth,cols=5,pos=h]
\caption{Performance comparison of SRA3, SRA and Two$\_$Arch2 in terms of the average HV values on DTLZ problems. The best result is highlighted in boldface.}\label{tbl1}
\begin{tabular*}{\tblwidth}{@{} CCCCC@{} }
\toprule
Problem & $m$ & SRA3 & SRA & Two$\_$Arch2 \\
\midrule
DTLZ1 & 5 & 9.74$e-$01 & \textbf{9.76$e-$01} & 9.76$e-$01\\
DTLZ1 & 10 & 9.99$e-$01 & \textbf{9.99$e-$01} & 9.95$e-$01\\
DTLZ1 & 15 & \textbf{9.98$e-$01} & 9.97$e-$01 & 9.86$e-$01\\
%\specialrule{0em}{0pt}{0.2pt}
\hdashline
% \specialrule{0em}{0.5pt}{0.5pt}
DTLZ2 & 5 & 7.77$e-$01 & \textbf{8.08$e-$01} & 7.67$e-$01\\
DTLZ2 & 10 & 9.38$e-$01 & \textbf{9.47$e-$01} & 7.40$e-$01\\
DTLZ2 & 15 & \textbf{9.34$e-$01} & 9.33$e-$01 & 6.11$e-$01\\
%\specialrule{0em}{0pt}{0.2pt}
\hdashline \specialrule{0em}{0pt}{0.02cm}
%\specialrule{0em}{0pt}{0.2pt}
DTLZ3 & 5 & \textbf{7.71$e-$01} & 7.71$e-$01 & 7.47$e-$01\\
DTLZ3 & 10 & 9.13$e-$01 & \textbf{9.30$e-$01} & 6.10$e-$01\\
DTLZ3 & 15 & \textbf{8.90$e-$01} & 8.81$e-$01 & 0.00$e+$00\\
%\specialrule{0em}{0pt}{0.2pt}
\hdashline \specialrule{0em}{0pt}{0.02cm}
%\specialrule{0em}{0pt}{0.2pt}
DTLZ4 & 5 & 7.87$e-$01 & \textbf{8.11$e-$01} & 7.54$e-$01\\
DTLZ4 & 10 & 9.53$e-$01 & \textbf{9.61$e-$01} & 7.18$e-$01\\
DTLZ4 & 15 & \textbf{9.63$e-$01} & 9.63$e-$01 & 6.33$e-$01\\
\bottomrule
\end{tabular*}
\end{table}

\begin{table}[width=.9\linewidth,cols=5,pos=h]
\caption{Performance comparison of SRA3, SRA and Two$\_$Arch2 in terms of the average IGD values on DTLZ problems. The best result is highlighted in boldface.}\label{tbl1}
\begin{tabular*}{\tblwidth}{@{} CCCCC@{} }
\toprule
Problem & $m$ & SRA3 & SRA & Two$\_$Arch2 \\
\midrule
DTLZ1 & 5 & 1.09$e-$01 & 1.02$e-$01 & \textbf{9.64$e-$02}\\
DTLZ1 & 10 & 1.96$e-$01 & 1.91$e-$01 & \textbf{1.87$e-$01}\\
DTLZ1 & 15 & 2.56$e-$01 & \textbf{2.47$e-$01} & 2.76$e-$01\\
%\specialrule{0em}{0pt}{0.2pt}
\hdashline \specialrule{0em}{0pt}{0.02cm}
%\specialrule{0em}{0pt}{0.2pt}
DTLZ2 & 5 & 1.80$e-$01 & 1.76$e-$01 & \textbf{1.52$e-$01}\\
DTLZ2 & 10 & 3.82$e-$01 & \textbf{3.67$e-$01} & 3.95$e-$01\\
DTLZ2 & 15 & 5.79$e-$01 & \textbf{5.39$e-$01} & 5.63$e-$01\\
%\specialrule{0em}{0pt}{0.2pt}
\hdashline \specialrule{0em}{0pt}{0.02cm}
%\specialrule{0em}{0pt}{0.2pt}
DTLZ3 & 5 & \textbf{1.78$e-$01} & 2.01$e-$01 & 1.89$e-$01\\
DTLZ3 & 10 & 4.04$e-$01 & \textbf{3.83$e-$01} & 6.22$e-$01\\
DTLZ3 & 15 & 6.04$e-$01 & \textbf{5.76$e-$01} & 2.28$e+$01\\
%\specialrule{0em}{0pt}{0.2pt}
\hdashline \specialrule{0em}{0pt}{0.02cm}
%\specialrule{0em}{0pt}{0.2pt}
DTLZ4 & 5 & 1.79$e-$01 & 1.78$e-$01 & \textbf{1.53$e-$01}\\
DTLZ4 & 10 & 3.86$e-$01 & \textbf{3.77$e-$01} & 3.87$e-$01\\
DTLZ4 & 15 & 5.73$e-$01 & 5.65$e-$01 & \textbf{5.51$e-$01}\\
\bottomrule
\end{tabular*}
\end{table}

\begin{table}[width=.9\linewidth,cols=5,pos=h]
\caption{Performance comparison of SRA3, SRA and Two$\_$Arch2 in terms of the average HV values on WFG problems. The best result is highlighted in boldface.}\label{tbl1}
\begin{tabular*}{\tblwidth}{@{} CCCCC@{} }
\toprule
Problem & $m$ & SRA3 & SRA & Two$\_$Arch2 \\
\midrule
WFG1 & 5 & 9.88$e-$01 & 9.47$e-$01 & \textbf{9.96$e-$01}\\
WFG1 & 10 & 9.50$e-$01 & 9.27$e-$01 & \textbf{9.97$e-$01}\\
WFG1 & 15 & 9.89$e-$01 & 8.93$e-$01 & \textbf{9.98$e-$01}\\
%\specialrule{0em}{0pt}{0.2pt}
\hdashline \specialrule{0em}{0pt}{0.02cm}
%\specialrule{0em}{0pt}{0.2pt}
WFG2 & 5 & 9.83$e-$01 & 9.79$e-$01 & \textbf{9.95$e-$01}\\
WFG2 & 10 & 9.89$e-$01 & 9.86$e-$01 & \textbf{9.95$e-$01}\\
WFG2 & 15 & 9.89$e-$01 & 9.88$e-$01 & \textbf{9.97$e-$01}\\
%\specialrule{0em}{0pt}{0.2pt}
\hdashline \specialrule{0em}{0pt}{0.02cm}
%\specialrule{0em}{0pt}{0.2pt}
WFG3 & 5 & 2.04$e-$02 & 3.11$e-$02 & \textbf{1.99$e-$01}\\
WFG3 & 10 & 0.00$e+$00 & 0.00$e+$00 & \textbf{4.12$e-$04}\\
WFG3 & 15 & 0.00$e+$00 & 0.00$e+$00 & 0.00$e+$00\\
%\specialrule{0em}{0pt}{0.2pt}
\hdashline \specialrule{0em}{0pt}{0.02cm}
%\specialrule{0em}{0pt}{0.2pt}
WFG4 & 5 & 6.98$e-$01 & 7.43$e-$01 & \textbf{7.53$e-$01}\\
WFG4 & 10 & 7.75$e-$01 & \textbf{8.49$e-$01} & 8.10$e-$01\\
WFG4 & 15 & 7.34$e-$01 & 7.98$e-$01 & \textbf{8.08$e-$01}\\
%\specialrule{0em}{0pt}{0.2pt}
\hdashline \specialrule{0em}{0pt}{0.02cm}
%\specialrule{0em}{0pt}{0.2pt}
WFG5 & 5 & 6.74$e-$01 & 7.07$e-$01 & \textbf{7.10$e-$01}\\
WFG5 & 10 & 7.64$e-$01 & \textbf{8.06$e-$01} & 7.64$e-$01\\
WFG5 & 15 & 7.52$e-$01 & \textbf{7.79$e-$01} & 6.87$e-$01\\
%\specialrule{0em}{0pt}{0.2pt}
\hdashline \specialrule{0em}{0pt}{0.02cm}
%\specialrule{0em}{0pt}{0.2pt}
WFG6 & 5 & 6.24$e-$01 & 6.89$e-$01 & \textbf{6.99$e-$01}\\
WFG6 & 10 & 6.95$e-$01 & \textbf{7.68$e-$01} & 7.39$e-$01\\
WFG6 & 15 & 6.58$e-$01 & \textbf{7.45$e-$01} & 7.26$e-$01\\
%\specialrule{0em}{0pt}{0.2pt}
\hdashline \specialrule{0em}{0pt}{0.02cm}
%\specialrule{0em}{0pt}{0.2pt}
WFG7 & 5 & 7.08$e-$01 & 7.62$e-$01 & \textbf{7.70$e-$01}\\
WFG7 & 10 & 8.28$e-$01 & \textbf{8.71$e-$01} & 8.35$e-$01\\
WFG7 & 15 & \textbf{8.36$e-$01} & 8.02$e-$01 & 7.90$e-$01\\
%\specialrule{0em}{0pt}{0.2pt}
\hdashline \specialrule{0em}{0pt}{0.02cm}
%\specialrule{0em}{0pt}{0.2pt}
WFG8 & 5 & 5.29$e-$01 & \textbf{6.38$e-$01} & 6.36$e-$01\\
WFG8 & 10 & 5.46$e-$01 & \textbf{7.28$e-$01} & 6.11$e-$01\\
WFG8 & 15 & 4.52$e-$01 & \textbf{6.21$e-$01} & 5.62$e-$01\\
%\specialrule{0em}{0pt}{0.2pt}
\hdashline \specialrule{0em}{0pt}{0.02cm}
%\specialrule{0em}{0pt}{0.2pt}
WFG9 & 5 & 7.07$e-$01 & \textbf{7.33$e-$01} & 7.20$e-$01\\
WFG9 & 10 & 7.62$e-$01 & \textbf{8.22$e-$01} & 6.86$e-$01\\
WFG9 & 15 & 6.79$e-$01 & \textbf{7.11$e-$01} & 6.14$e-$01\\

\bottomrule
\end{tabular*}
\end{table}

\begin{table}[width=.9\linewidth,cols=5,pos=h]
\caption{Performance comparison of SRA3, SRA and Two$\_$Arch2 in terms of the average IGD values on WFG problems. The best result is highlighted in boldface.}\label{tbl1}
\begin{tabular*}{\tblwidth}{@{} CCCCC@{} }
\toprule
Problem & $m$ & SRA3 & SRA & Two$\_$Arch2 \\
\midrule
WFG1 & 5 & 8.02$e-$02 & 7.20$e-$02 & \textbf{5.27$e-$02}\\
WFG1 & 10 & 1.10$e-$01 & 1.05$e-$01 & \textbf{7.74$e-$02}\\
WFG1 & 15 & 1.15$e-$01 & 1.42$e-$01 & \textbf{8.35$e-$02}\\
%\specialrule{0em}{0pt}{0.2pt}
\hdashline \specialrule{0em}{0pt}{0.02cm}
%\specialrule{0em}{0pt}{0.2pt}
WFG2 & 5 & 6.58$e-$02 & 6.64$e-$02 & \textbf{5.76$e-$02}\\
WFG2 & 10 & 8.55$e-$02 & 8.75$e-$02 & \textbf{8.15$e-$02}\\
WFG2 & 15 & 9.44$e-$02 & 1.07$e-$01 & \textbf{8.65$e-$02}\\
%\specialrule{0em}{0pt}{0.2pt}
\hdashline \specialrule{0em}{0pt}{0.02cm}
%\specialrule{0em}{0pt}{0.2pt}
WFG3 & 5 & 6.68$e-$01 & 5.24$e-$01 & \textbf{1.36$e-$01}\\
WFG3 & 10 & 2.62$e+$01 & 1.71$e+$01 & \textbf{2.49$e-$00}\\
WFG3 & 15 & 2.56$e+$03 & 1.37$e+$03 & \textbf{6.51$e+$01}\\
%\specialrule{0em}{0pt}{0.2pt}
\hdashline \specialrule{0em}{0pt}{0.02cm}
%\specialrule{0em}{0pt}{0.2pt}
WFG4 & 5 & 1.73$e-$01 & 1.72$e-$01 & \textbf{1.57$e-$01}\\
WFG4 & 10 & 3.85$e-$01 & 4.27$e-$01 & \textbf{3.84$e-$01}\\
WFG4 & 15 & 5.54$e-$01 & 6.06$e-$01 & \textbf{5.17$e-$01}\\
%\specialrule{0em}{0pt}{0.2pt}
\hdashline \specialrule{0em}{0pt}{0.02cm}
%\specialrule{0em}{0pt}{0.2pt}
WFG5 & 5 & 1.78$e-$01 & 1.70$e-$01 & \textbf{1.57$e-$01}\\
WFG5 & 10 & 3.90$e-$01 & 4.05$e-$01 & \textbf{3.79$e-$01}\\
WFG5 & 15 & 5.41$e-$01 & 6.35$e-$01 & \textbf{5.12$e-$01}\\
%\specialrule{0em}{0pt}{0.2pt}
\hdashline \specialrule{0em}{0pt}{0.02cm}
%\specialrule{0em}{0pt}{0.2pt}
WFG6 & 5 & 1.92$e-$01 & 1.77$e-$01 & \textbf{1.63$e-$01}\\
WFG6 & 10 & 4.03$e-$01 & 4.18$e-$01 & \textbf{3.93$e-$01}\\
WFG6 & 15 & 5.97$e-$01 & 6.44$e-$01 & \textbf{5.20$e-$01}\\
%\specialrule{0em}{0pt}{0.2pt}
\hdashline \specialrule{0em}{0pt}{0.02cm}
%\specialrule{0em}{0pt}{0.2pt}
WFG7 & 5 & 1.82$e-$01 & 1.77$e-$01 & \textbf{1.55$e-$01}\\
WFG7 & 10 & 3.94$e-$01 & 4.11$e-$01 & \textbf{3.79$e-$01}\\
WFG7 & 15 & 5.66$e-$01 & 5.96$e-$01 & \textbf{5.15$e-$01}\\
%\specialrule{0em}{0pt}{0.2pt}
\hdashline \specialrule{0em}{0pt}{0.02cm}
%\specialrule{0em}{0pt}{0.2pt}
WFG8 & 5 & 2.25$e-$01 & \textbf{1.96$e-$01} & 1.97$e-$01\\
WFG8 & 10 & 4.65$e-$01 & \textbf{4.47$e-$01} & 4.65$e-$01\\
WFG8 & 15 & 6.14$e-$01 & 5.96$e-$01 & \textbf{5.92$e-$01}\\
%\specialrule{0em}{0pt}{0.2pt}
\hdashline \specialrule{0em}{0pt}{0.02cm}
%\specialrule{0em}{0pt}{0.2pt}
WFG9 & 5 & 1.67$e-$01 & 1.66$e-$01 & \textbf{1.53$e-$01}\\
WFG9 & 10 & \textbf{3.86$e-$01} & 4.07$e-$01 & 3.92$e-$01\\
WFG9 & 15 & 5.45$e-$01 & 5.77$e-$01 & \textbf{5.38$e-$01}\\

\bottomrule
\end{tabular*}
\end{table}

The experimental results show that SRA3 performs  similarly to SRA and better than Two$\_$Arch2 on the DTLZ problems, especially on the DTLZ3 problem, where Two$\_$Arch2 has difficulty converging on the 15-objective DTLZ3 problem. But SRA3 performs poorly on WFG, a problem set where the range of each objective is different. And it is pointed out in \cite{li2019quality} that when the objectives are incomparable, usually the larger the range of the objective, the larger the impact of that objective on the indicator value, contrary to the fact that we want all objectives to contribute equally to the indicator. Meanwhile, \cite{zitzler2004indicator} points out that the value of indicator $I_{\epsilon+}(A,B)$ fluctuates a lot in different problems, which causes great trouble in determining the parameter $k$ of $I_{\epsilon+}$ indicator. Based on this, we tried to normalize SRA3, hoping to eliminate the bias caused by the different value ranges of different objectives and to keep the $I_{\epsilon+}$ indicator in a fixed range.

\subsection{Impact of normalization on algorithms}

In Section 3.3, we normalized SRA3 and compared its performance on each problem before and after normalization. Then we analyzed the effect of normalization on the IB-MOEAs, and finally we compared SRA3 with state-of-the-art algorithms on DTLZ and WFG problems.

\subsubsection{Normalization method}

For the CA archive, each objective is scaled to [0,1] before calculating the indicator value for each individual. Then, after calculating the $I_{\epsilon+}(x,y)$ for each individual, calculate the maximum absolute indicator value of $I_{\epsilon+}(x,y)$, scale the value of indicator $I_{\epsilon+}(x,y)$ to [-1,1] and calculate the $I_1(x)$ for all individuals. Finally, a new environment selection method for updating CA archive with reference to \cite{zitzler2004indicator} is used, and the update method of the normalized CA archive is shown in Algorithm 5. As for the DA archive, only the normalization of $H_2$ should be added in front of Algorithm 4, which will not be shown here.

\renewcommand{\algorithmicrequire}{\textbf{Input:}}  % Use Input in the format of Algorithm 2  
\renewcommand{\algorithmicensure}{\textbf{Output:}} % Use Output in the format of Algorithm 2
\begin{algorithm}
	\caption{\textit{updateCA($H_1$)} procedure (normalization)} 
	\begin{algorithmic}[1]
	\Require  $H_1$ (the union set of CA and $Q_t$)
    \Ensure  the updated CA
        \State $N\leftarrow{H_1}/2$
        \State for each objective $f_i$ calculate its upper bound $max_i=max_{x\in{H_1}}f_i(x)$ and its lower bound $min_i=min_{x\in{H_1}}f_i(x)$
        \State scale each objective to [0,1] by $f_i^{'}(x)=(f_i(x)-min_i)/(max_i-min_i)$
		\State calculate the indicator values $I_{\epsilon+}(x,y)$ using the scaled objective values $f_i^{'}(x)$ and calculate the maximum absolute indicator value $c=max_{x_1,x_2\in{H_1}}|I_{\epsilon+}(x_1,x_2)|$
		\State for $x_1\in{H_1}$, set $I_1(x_1)=\sum_{x_2\in{H_1\backslash{x_1}}}-e^{-I_{\epsilon+}(x_2,x_1)/(c\cdot{k})}$
		\While{$len(H_1)>N$}
		    \State choose an individual $x^*$ with the worst indicator value $I_1(x^*)$
		    \State $H_1\leftarrow{H_1}\backslash{x^*}$
		    \State $I_1(x)\leftarrow{I_1(x)+e^{-I_{\epsilon+}(x^*, x)/(c\cdot{k})}}$ for all $x\in{H_1}$
		\EndWhile
		\State \Return $H_1$
	\end{algorithmic} 
\end{algorithm}

After normalization, the computational complexity of CA archive update is $O(N^2)$, and the computational complexity of DA archive update remains $O(N)$. Compared with SRA, the environment selection of SRA3 still has a significant efficiency advantage in actual operation because SRA requires a large number of exchanges in environment selection using the stochastic ranking technique.

Meanwhile, in order to verify the effect of normalization, we compared SRA3 before and after normalization on DTLZ and WFG problems, evaluated them by HV and IGD indicators, and the results are shown in Tables 6 to 9.

\begin{table}[width=.9\linewidth,cols=4,pos=h]
\caption{Performance comparison of SRA3 before and after normalization through the average HV values on DTLZ problems. The best result is highlighted in boldface.}\label{tbl1}
\begin{tabular*}{\tblwidth}{@{} CCCCC@{} }
\toprule
Problem & $m$ & SRA3 & SRA3$_{norm}$  \\
\midrule
DTLZ1 & 5 & \textbf{9.74$e-$01} & 8.58$e-$01 \\
DTLZ1 & 10 & \textbf{9.99$e-$01} & 9.72$e-$01 \\
DTLZ1 & 15 & \textbf{9.98$e-$01} & 8.74$e-$01 \\
%\specialrule{0em}{0pt}{0.2pt}
\hdashline \specialrule{0em}{0pt}{0.02cm}
%\specialrule{0em}{0pt}{0.2pt}
DTLZ2 & 5 & 7.77$e-$01 & \textbf{8.05$e-$01} \\
DTLZ2 & 10 & 9.38$e-$01 & \textbf{9.73$e-$01} \\
DTLZ2 & 15 & 9.34$e-$01 & \textbf{9.85$e-$01} \\
%\specialrule{0em}{0pt}{0.2pt}
\hdashline \specialrule{0em}{0pt}{0.02cm}
%\specialrule{0em}{0pt}{0.2pt}
DTLZ3 & 5 & \textbf{7.71$e-$01} & 3.80$e-$01 \\
DTLZ3 & 10 & \textbf{9.13$e-$01} & 6.15$e-$01 \\
DTLZ3 & 15 & \textbf{8.90$e-$01} & 7.07$e-$01 \\
%\specialrule{0em}{0pt}{0.2pt}
\hdashline \specialrule{0em}{0pt}{0.02cm}
%\specialrule{0em}{0pt}{0.2pt}
DTLZ4 & 5 & 7.87$e-$01 & \textbf{8.06$e-$01} \\
DTLZ4 & 10 & 9.53$e-$01 & \textbf{9.73$e-$01} \\
DTLZ4 & 15 & 9.63$e-$01 & \textbf{9.87$e-$01} \\
\bottomrule
\end{tabular*}
\end{table}

\begin{table}[width=.9\linewidth,cols=4,pos=h]
\caption{Performance comparison of SRA3 before and after normalization through the average IGD values on DTLZ problems. The best result is highlighted in boldface.}\label{tbl1}
\begin{tabular*}{\tblwidth}{@{} CCCCC@{} }
\toprule
Problem & $m$ & SRA3 & SRA3$_{norm}$ \\
\midrule
DTLZ1 & 5 & \textbf{1.09$e-$01} & 2.82$e-$01 \\
DTLZ1 & 10 & \textbf{1.96$e-$01} & 3.87$e-$01 \\
DTLZ1 & 15 & \textbf{2.56$e-$01} & 5.55$e-$01 \\
%\specialrule{0em}{0pt}{0.2pt}
\hdashline \specialrule{0em}{0pt}{0.02cm}
%\specialrule{0em}{0pt}{0.2pt}
DTLZ2 & 5 & 1.80$e-$01 & \textbf{1.73$e-$01} \\
DTLZ2 & 10 & \textbf{3.82$e-$01} & 3.85$e-$01 \\
DTLZ2 & 15 & 5.79$e-$01 & \textbf{5.52$e-$01} \\
%\specialrule{0em}{0pt}{0.2pt}
\hdashline \specialrule{0em}{0pt}{0.02cm}
%\specialrule{0em}{0pt}{0.2pt}
DTLZ3 & 5 & \textbf{1.78$e-$01} & 5.35$e-$01 \\
DTLZ3 & 10 & \textbf{4.04$e-$01} & 6.49$e-$01 \\
DTLZ3 & 15 & \textbf{6.04$e-$01} & 7.51$e-$01 \\
%\specialrule{0em}{0pt}{0.2pt}
\hdashline \specialrule{0em}{0pt}{0.02cm}
%\specialrule{0em}{0pt}{0.2pt}
DTLZ4 & 5 & 1.79$e-$01 & \textbf{1.71$e-$01} \\
DTLZ4 & 10 & 3.86$e-$01 & \textbf{3.83$e-$01} \\
DTLZ4 & 15 & 5.73$e-$01 & \textbf{5.53$e-$01} \\
\bottomrule
\end{tabular*}
\end{table}

\begin{table}[width=.9\linewidth,cols=4,pos=h]
\caption{Performance comparison of SRA3 before and after normalization through the average HV values on WFG problems. The best result is highlighted in boldface.}\label{tbl1}
\begin{tabular*}{\tblwidth}{@{} CCCCC@{} }
\toprule
Problem & $m$ & SRA3 & SRA3$_{norm}$ \\
\midrule
WFG1 & 5 & 9.88$e-$01 & \textbf{9.90$e-$01} \\
WFG1 & 10 & 9.50$e-$01 & \textbf{9.94$e-$01} \\
WFG1 & 15 & 9.89$e-$01 & \textbf{9.92$e-$01} \\
%\specialrule{0em}{0pt}{0.2pt}
\hdashline \specialrule{0em}{0pt}{0.02cm}
%\specialrule{0em}{0pt}{0.2pt}
WFG2 & 5 & 9.83$e-$01 & \textbf{9.88$e-$01} \\
WFG2 & 10 & 9.89$e-$01 & \textbf{9.93$e-$01} \\
WFG2 & 15 & 9.89$e-$01 & \textbf{9.91$e-$01} \\
%\specialrule{0em}{0pt}{0.2pt}
\hdashline \specialrule{0em}{0pt}{0.02cm}
%\specialrule{0em}{0pt}{0.2pt}
WFG3 & 5 & 2.04$e-$02 & \textbf{2.36$e-$01} \\
WFG3 & 10 & 0.00$e+$00 & 0.00$e+$00 \\
WFG3 & 15 & 0.00$e+$00 & 0.00$e+$00 \\
%\specialrule{0em}{0pt}{0.2pt}
\hdashline \specialrule{0em}{0pt}{0.02cm}
%\specialrule{0em}{0pt}{0.2pt}
WFG4 & 5 & 6.98$e-$01 & \textbf{7.83$e-$01} \\
WFG4 & 10 & 7.75$e-$01 & \textbf{9.23$e-$01} \\
WFG4 & 15 & 7.34$e-$01 & \textbf{8.83$e-$01} \\
%\specialrule{0em}{0pt}{0.2pt}
\hdashline \specialrule{0em}{0pt}{0.02cm}
%\specialrule{0em}{0pt}{0.2pt}
WFG5 & 5 & 6.74$e-$01 & \textbf{7.42$e-$01} \\
WFG5 & 10 & 7.64$e-$01 & \textbf{8.84$e-$01} \\
WFG5 & 15 & 7.52$e-$01 & \textbf{8.62$e-$01} \\
%\specialrule{0em}{0pt}{0.2pt}
\hdashline \specialrule{0em}{0pt}{0.02cm}
%\specialrule{0em}{0pt}{0.2pt}
WFG6 & 5 & 6.24$e-$01 & \textbf{7.31$e-$01} \\
WFG6 & 10 & 6.95$e-$01 & \textbf{8.65$e-$01} \\
WFG6 & 15 & 6.58$e-$01 & \textbf{8.59$e-$01} \\
%\specialrule{0em}{0pt}{0.2pt}
\hdashline \specialrule{0em}{0pt}{0.02cm}
%\specialrule{0em}{0pt}{0.2pt}
WFG7 & 5 & 7.08$e-$01 & \textbf{7.96$e-$01} \\
WFG7 & 10 & 8.28$e-$01 & \textbf{9.53$e-$01} \\
WFG7 & 15 & 8.36$e-$01 & \textbf{9.72$e-$01} \\
%\specialrule{0em}{0pt}{0.2pt}
\hdashline \specialrule{0em}{0pt}{0.02cm}
%\specialrule{0em}{0pt}{0.2pt}
WFG8 & 5 & 5.29$e-$01 & \textbf{6.77$e-$01} \\
WFG8 & 10 & 5.46$e-$01 & \textbf{8.64$e-$01} \\
WFG8 & 15 & 4.52$e-$01 & \textbf{8.33$e-$01} \\
%\specialrule{0em}{0pt}{0.2pt}
\hdashline \specialrule{0em}{0pt}{0.02cm}
%\specialrule{0em}{0pt}{0.2pt}
WFG9 & 5 & 7.07$e-$01 & \textbf{7.34$e-$01} \\
WFG9 & 10 & 7.62$e-$01 & \textbf{8.25$e-$01} \\
WFG9 & 15 & 6.79$e-$01 & \textbf{7.60$e-$01} \\

\bottomrule
\end{tabular*}
\end{table}

\begin{table}[width=.9\linewidth,cols=5,pos=h]
\caption{Performance comparison of SRA3 before and after normalization through the average IGD values on WFG problems. The best result is highlighted in boldface.}\label{tbl1}
\begin{tabular*}{\tblwidth}{@{} CCCCC@{} }
\toprule
Problem & $m$ & SRA3 & SRA3$_{norm}$ \\
\midrule
WFG1 & 5 & 8.02$e-$02 & \textbf{5.77$e-$02} \\
WFG1 & 10 & 1.10$e-$01 & \textbf{8.49$e-$02} \\
WFG1 & 15 & 1.15$e-$01 & \textbf{1.05$e-$01} \\
%\specialrule{0em}{0pt}{0.2pt}
\hdashline \specialrule{0em}{0pt}{0.02cm}
%\specialrule{0em}{0pt}{0.2pt}
WFG2 & 5 & 6.58$e-$02 & \textbf{6.49$e-$02} \\
WFG2 & 10 & \textbf{8.55$e-$02} & 9.17$e-$02 \\
WFG2 & 15 & \textbf{9.44$e-$02} & 1.12$e-$01 \\
%\specialrule{0em}{0pt}{0.2pt}
\hdashline \specialrule{0em}{0pt}{0.02cm}
%\specialrule{0em}{0pt}{0.2pt}
WFG3 & 5 & 6.68$e-$01 & \textbf{8.65$e-$02} \\
WFG3 & 10 & 2.62$e+$01 & \textbf{4.32$e+$00} \\
WFG3 & 15 & 2.56$e+$03 & \textbf{1.30$e+$02} \\
%\specialrule{0em}{0pt}{0.2pt}
\hdashline \specialrule{0em}{0pt}{0.02cm}
%\specialrule{0em}{0pt}{0.2pt}
WFG4 & 5 & 1.73$e-$01 & \textbf{1.69$e-$01} \\
WFG4 & 10 & 3.85$e-$01 & \textbf{3.65$e-$01} \\
WFG4 & 15 & 5.54$e-$01 & \textbf{5.38$e-$01} \\
%\specialrule{0em}{0pt}{0.2pt}
\hdashline \specialrule{0em}{0pt}{0.02cm}
%\specialrule{0em}{0pt}{0.2pt}
WFG5 & 5 & 1.78$e-$01 & \textbf{1.68$e-$01} \\
WFG5 & 10 & 3.90$e-$01 & \textbf{3.63$e-$01} \\
WFG5 & 15 & 5.41$e-$01 & \textbf{5.33$e-$01} \\
%\specialrule{0em}{0pt}{0.2pt}
\hdashline \specialrule{0em}{0pt}{0.02cm}
%\specialrule{0em}{0pt}{0.2pt}
WFG6 & 5 & 1.92$e-$01 & \textbf{1.72$e-$01} \\
WFG6 & 10 & 4.03$e-$01 & \textbf{3.72$e-$01} \\
WFG6 & 15 & 5.97$e-$01 & \textbf{5.66$e-$01} \\
%\specialrule{0em}{0pt}{0.2pt}
\hdashline \specialrule{0em}{0pt}{0.02cm}
%\specialrule{0em}{0pt}{0.2pt}
WFG7 & 5 & 1.82$e-$01 & \textbf{1.73$e-$01} \\
WFG7 & 10 & 3.94$e-$01 & \textbf{3.72$e-$01} \\
WFG7 & 15 & 5.66$e-$01 & \textbf{5.57$e-$01} \\
%\specialrule{0em}{0pt}{0.2pt}
\hdashline \specialrule{0em}{0pt}{0.02cm}
%\specialrule{0em}{0pt}{0.2pt}
WFG8 & 5 & 2.25$e-$01 & \textbf{1.92$e-$01} \\
WFG8 & 10 & 4.65$e-$01 & \textbf{3.73$e-$01} \\
WFG8 & 15 & 6.14$e-$01 & \textbf{5.79$e-$01} \\
%\specialrule{0em}{0pt}{0.2pt}
\hdashline \specialrule{0em}{0pt}{0.02cm}
%\specialrule{0em}{0pt}{0.2pt}
WFG9 & 5 & 1.67$e-$01 & \textbf{1.58$e-$01} \\
WFG9 & 10 & 3.86$e-$01 & \textbf{3.47$e-$01} \\
WFG9 & 15 & 5.45$e-$01 & \textbf{5.03$e-$01} \\

\bottomrule
\end{tabular*}
\end{table}

Experimental results show that SRA3 performs better after normalization for most problems, especially on WFG, a problem set with different ranges of each objective, normalization makes each objective have the same contribution to the indicators, resulting in a significant improvement in SRA3 performance. However, on the DTLZ1 and DTLZ3 problems, the performance of SRA3 becomes worse after normalization.

\subsubsection{normalized analysis}

Several IB-MOEAs (SRA3, SRA, IBEA) perform better after normalization on most problems, but worse on the DTLZ1 and DTLZ3 problems (comparison results before and after normalization for SRA and IBEA are not shown due to space limitations). Therefore, in order to explore the impact of normalization on indicator-based algorithms, we divided the SRA3 algorithm into four versions: without normalization, normalizing only the $I_{\epsilon+}$ indicator, normalizing only the $I_{SDE}$ indicator, and normalizing both indicators, and then conducted comparative experiments on the 15-objective DTLZ and WFG problems. Among them, for the DTLZ problems, we compared the performance of the four versions of SRA3 on HV and IGD indicators, and the results are shown in Tables 10 and 11. The parallel coordinate plots of DLTZ2 and DTLZ3 problems for 15 objectives were further analyzed, and the results are shown in Figures 1 and 2. As for the WFG problems, normalization eliminates the effect of different objectives contributing inconsistently to the indicators due to the inconsistent range of each objective, but this also results in the impact of normalization in other aspects not being reflected in the HV and IGD indicators. Therefore, we directly analyzed the parallel coordinate plots of WFG4 and WFG6 problems for 15 objectives, and the results are shown in Figures 3 and 4.

\begin{table}[width=.9\linewidth,cols=5,pos=h]
\caption{Performance comparison of the four versions of SRA3 in terms of the average HV values on the 15-objective DTLZ problems. The best result is highlighted in boldface.}\label{tbl1}
\begin{tabular*}{\tblwidth}{@{} CCCCC@{} }
\toprule
Problem & SRA3 & SRA3$_{normI_{\epsilon}}$ & SRA3$_{normSDE}$ & SRA3$_{normAll}$ \\
\midrule
DTLZ1 & \textbf{9.98$e-$01} & 9.91$e-$1 & 9.93$e-$1 & 8.74$e-$1\\
DTLZ2 & 9.34$e-$01 & \textbf{9.86$e-$1} & 9.24$e-$1 & 9.85$e-$1\\
DTLZ3 & \textbf{8.90$e-$01} & 7.44$e-$1 & 8.61$e-$1 & 7.07$e-$1\\
DTLZ4 & 9.63$e-$01 & \textbf{9.88$e-$1} & 9.60$e-$1 & 9.87$e-$1\\
\bottomrule
\end{tabular*}
\end{table}

\begin{table}[width=.9\linewidth,cols=4,pos=h]
\caption{Performance comparison of the four versions of SRA3 in terms of the average IGD values on the 15-objective DTLZ problems. The best result is highlighted in boldface.}\label{tbl1}
\begin{tabular*}{\tblwidth}{@{} CCCCC@{} }
\toprule
Problem & SRA3 & SRA3$_{normI_{\epsilon}}$ & SRA3$_{normSDE}$ & SRA3$_{normAll}$ \\
\midrule
DTLZ1 & \textbf{2.56$e-$01} & 4.43$e-$01 & 2.56$e-$01 & 5.55$e-$01 \\
DTLZ2 & 5.79$e-$01 & \textbf{5.52$e-$01} & 5.64$e-$01 & 5.52$e-$01 \\
DTLZ3 & \textbf{6.04$e-$01} & 7.24$e-$01 & 6.08$e-$01 & 7.51$e-$01 \\
DTLZ4 & 5.73$e-$01 & 5.52$e-$01 & \textbf{5.52$e-$01} & 5.53$e-$01 \\
\bottomrule
\end{tabular*}
\end{table}

\begin{figure*}
    \centering
    \begin{subfigure}{4cm}
        \includegraphics[width=4cm]{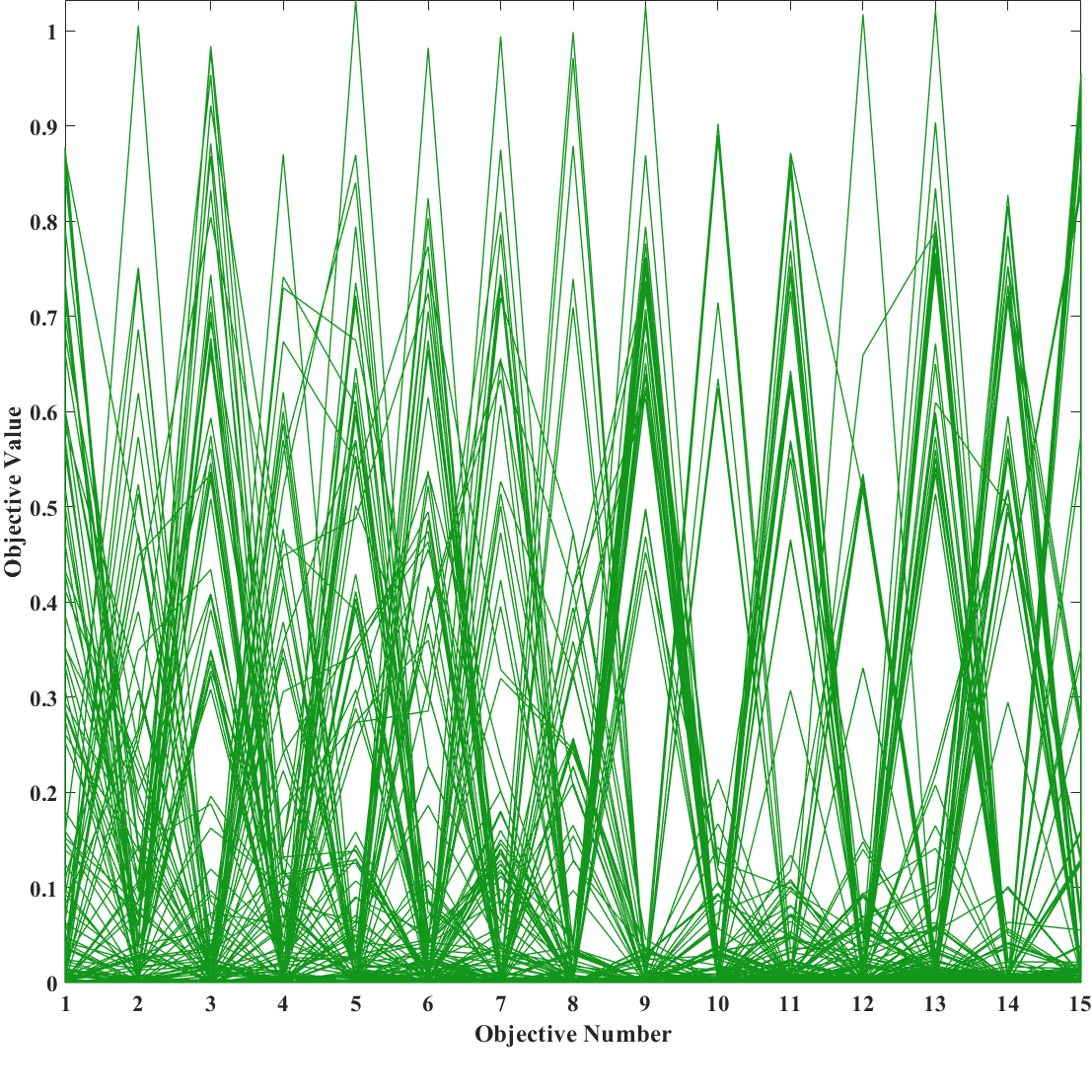}
        \subcaption*{SRA3 without normalized}
    \end{subfigure}
    \begin{subfigure}{4cm}
        \includegraphics[width=4cm]{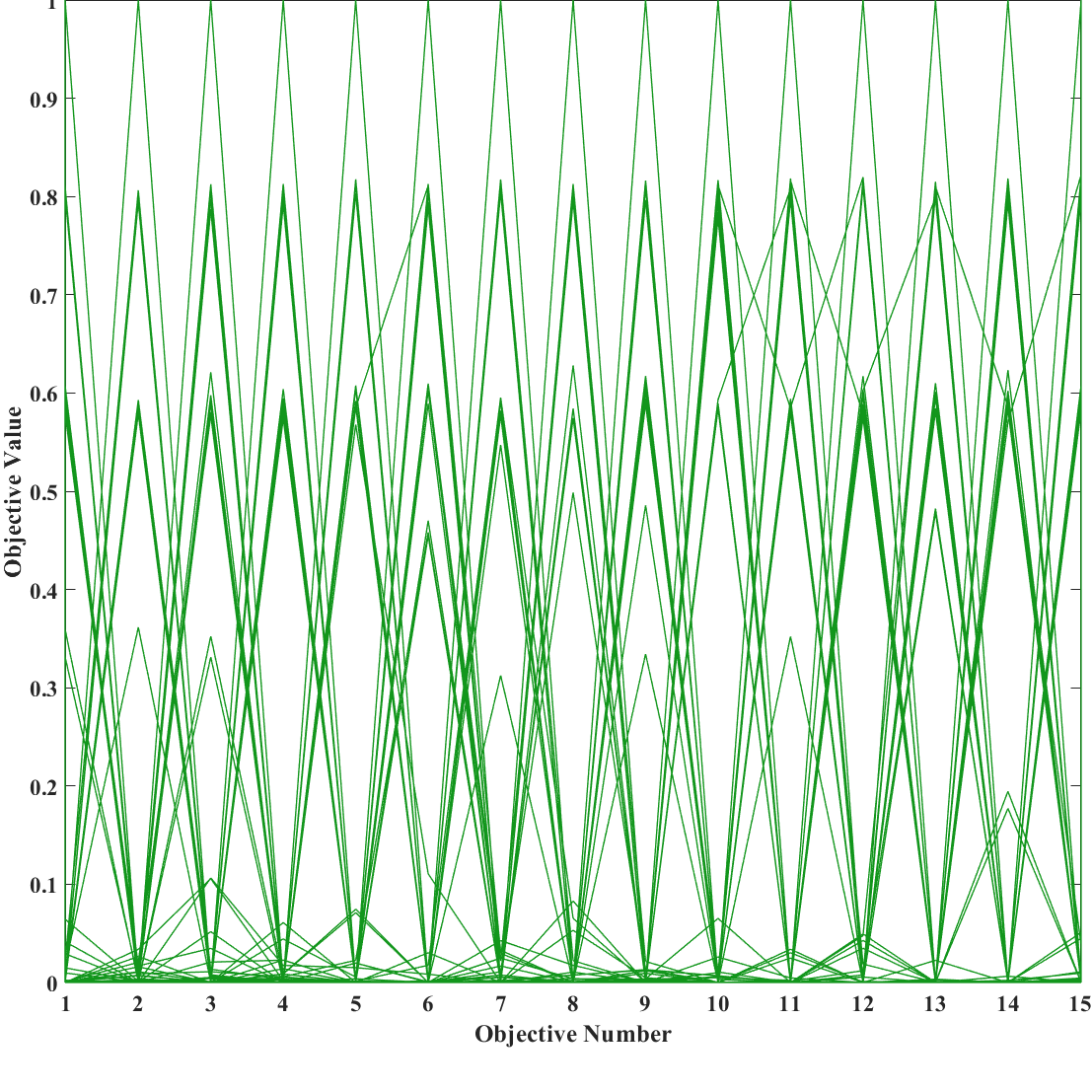}
        \subcaption*{SRA3 for normalized $I_{\epsilon+}$}
    \end{subfigure}
    \begin{subfigure}{4cm}
        \includegraphics[width=4cm]{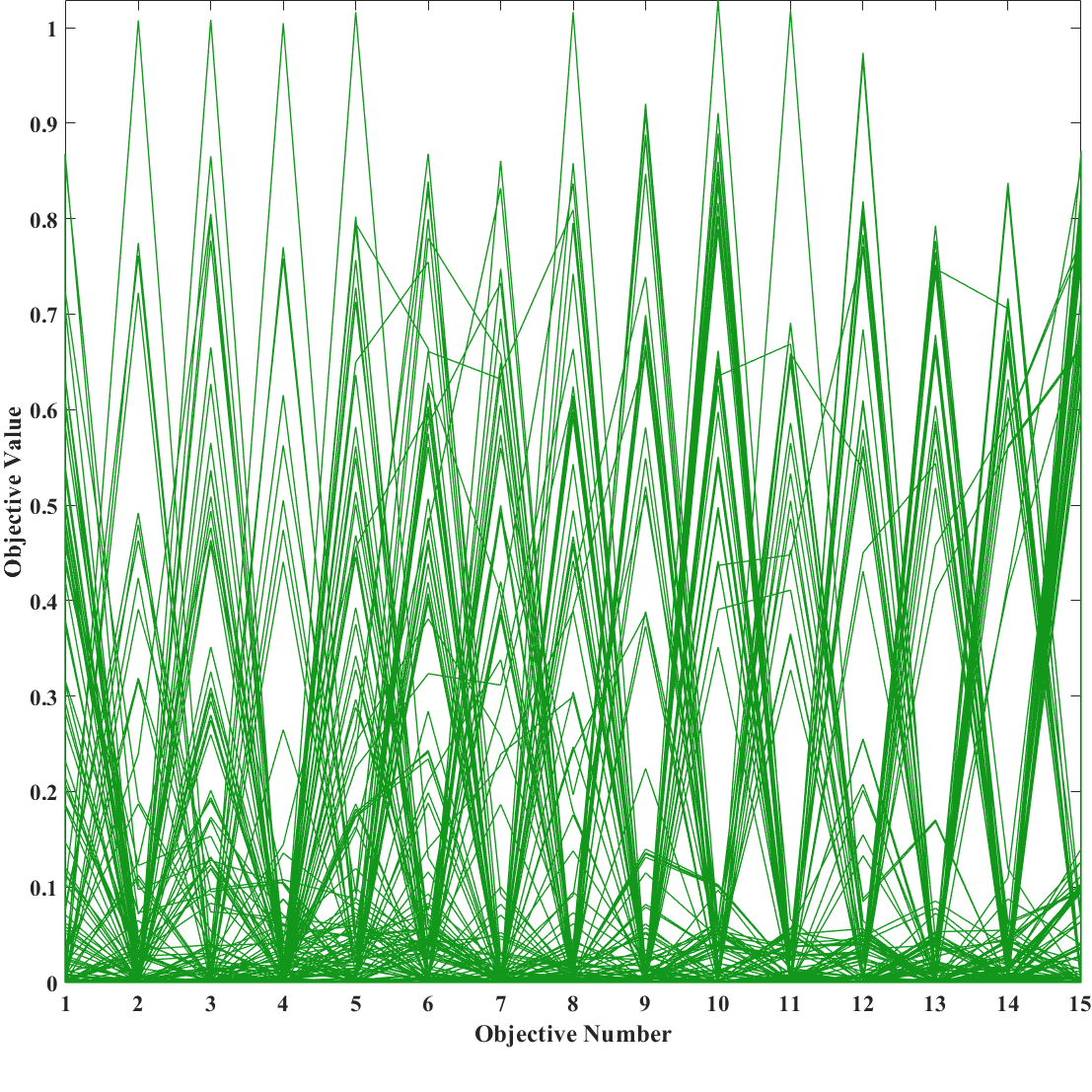}
        \subcaption*{SRA3 for normalized $I_{SDE}$}
    \end{subfigure}
    \begin{subfigure}{4cm}
        \includegraphics[width=4cm]{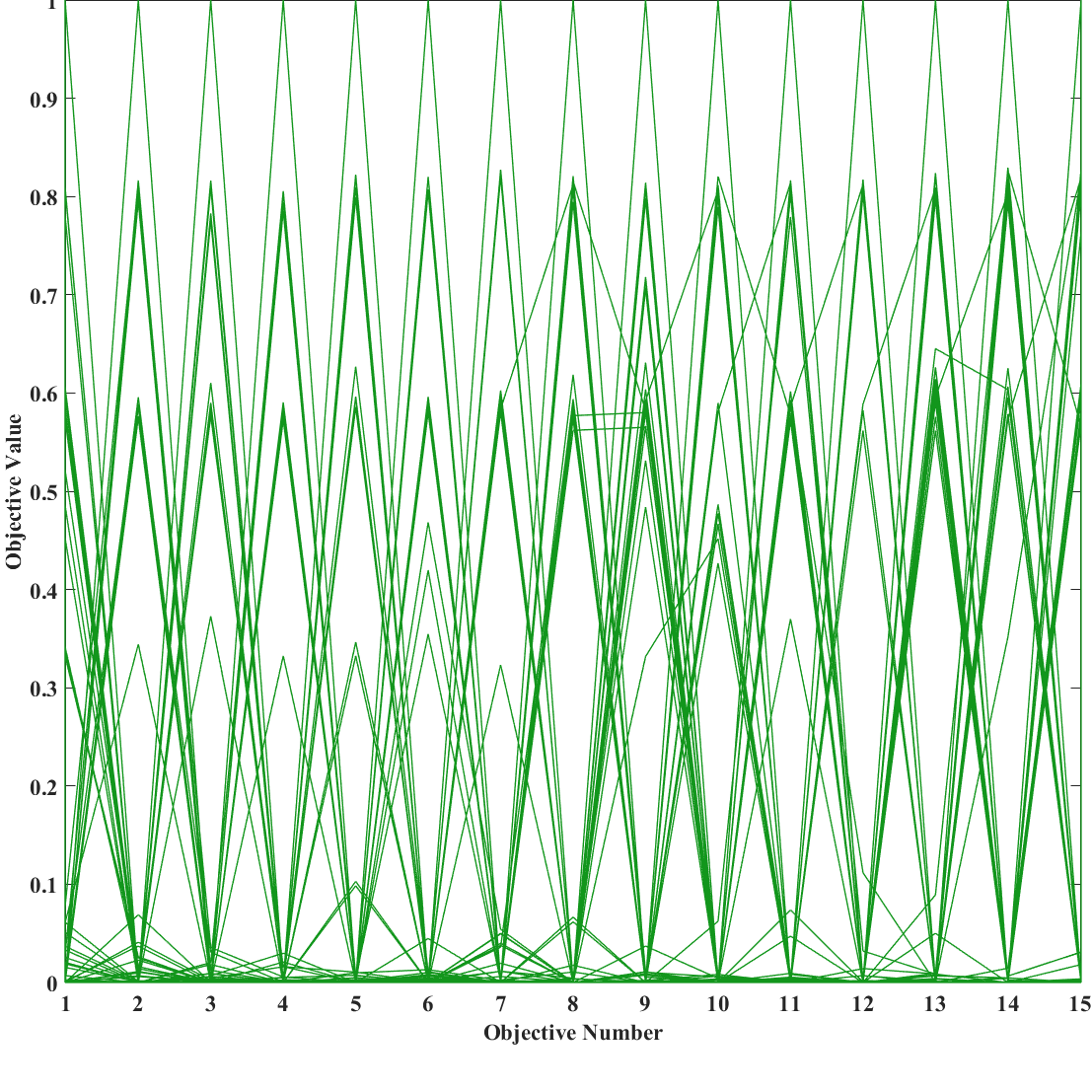}
        \subcaption*{Both normalized SRA3}
    \end{subfigure}
    \caption{Comparison of parallel coordinate plots of four versions of SRA3 on the DTLZ2 problem with 15 objectives.}
\end{figure*}

\begin{figure*}
    \centering
    \begin{subfigure}{4cm}
        \includegraphics[width=4cm]{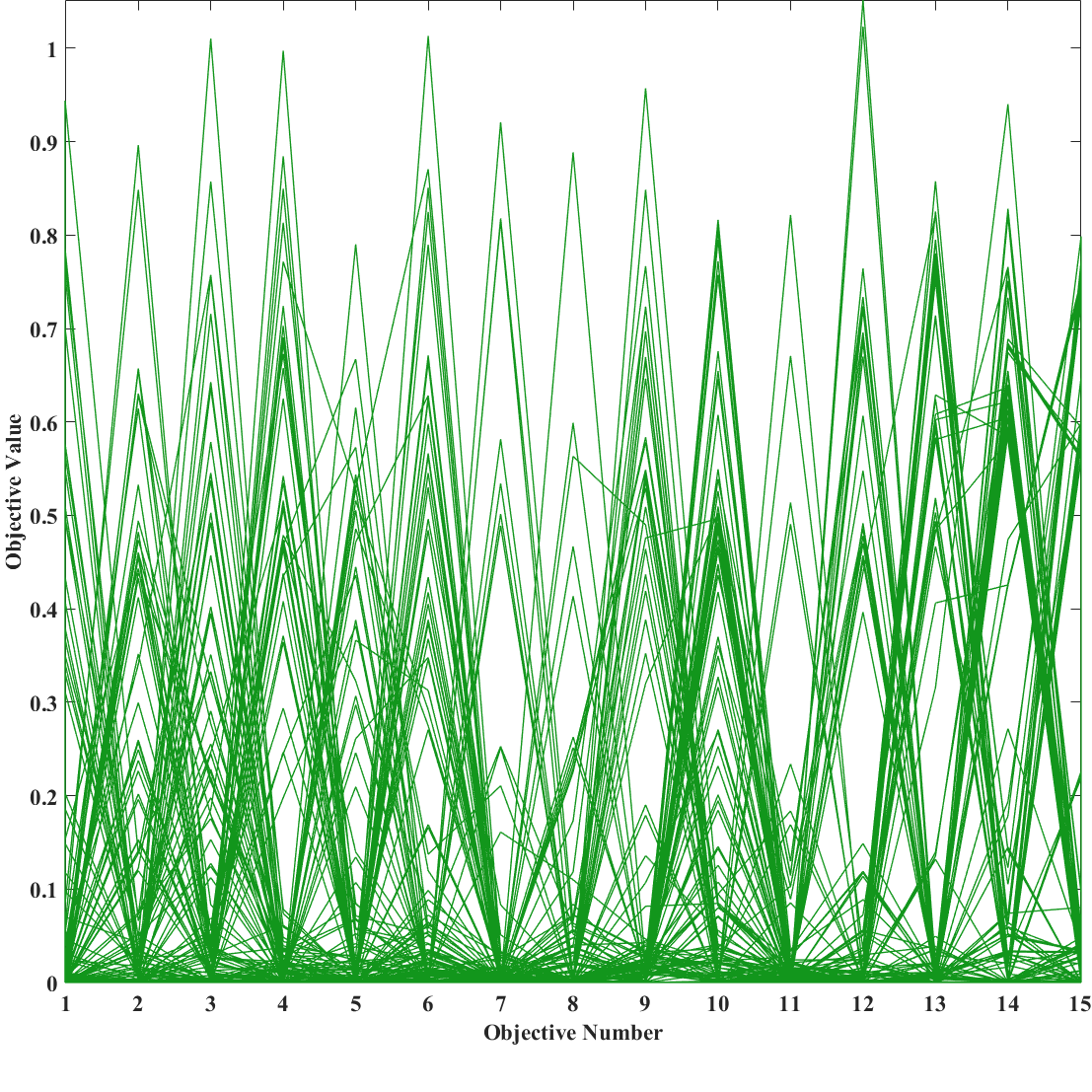}
        \subcaption*{SRA3 without normalized}
    \end{subfigure}
    \begin{subfigure}{4cm}
        \includegraphics[width=4cm]{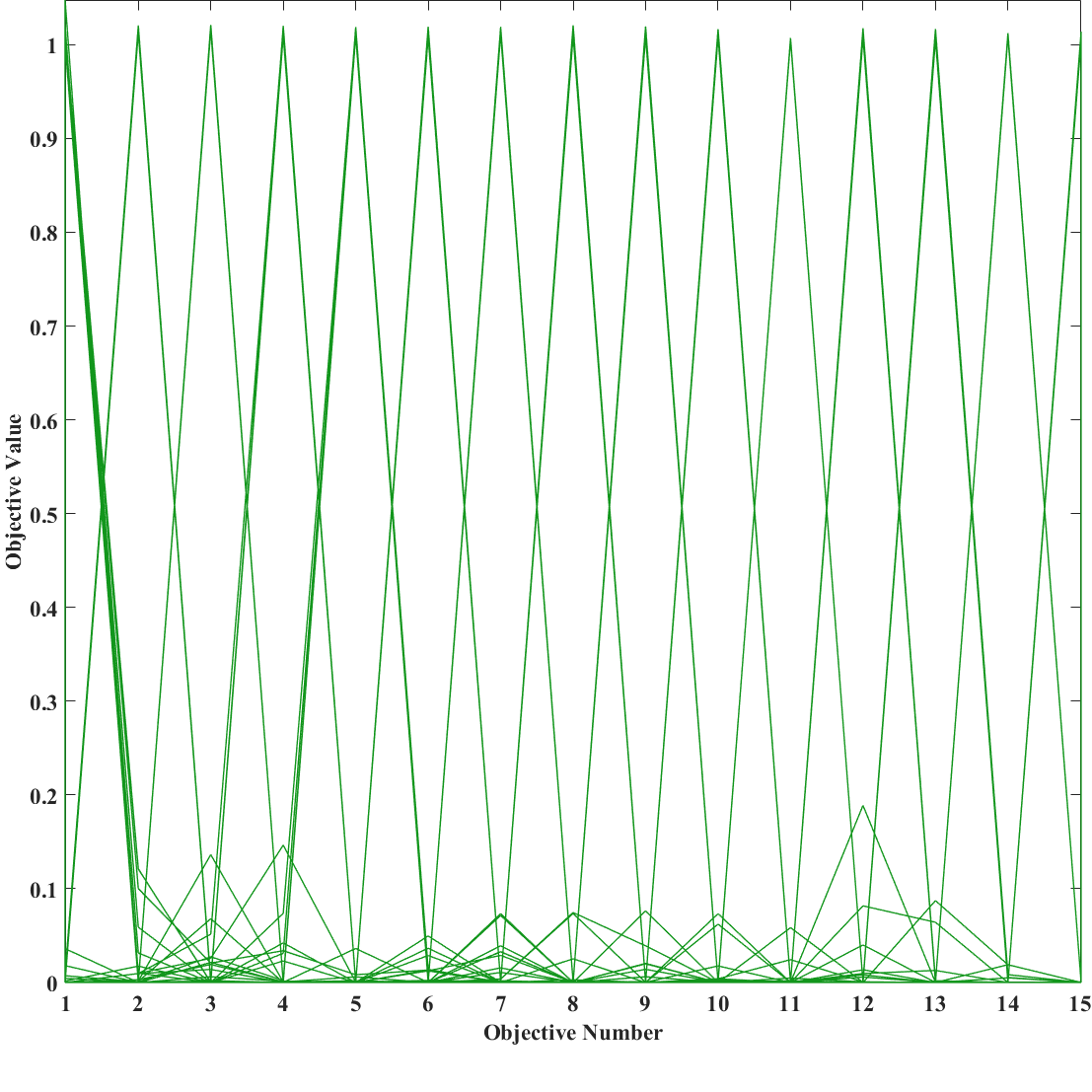}
        \subcaption*{SRA3 for normalized $I_{\epsilon+}$}
    \end{subfigure}
    \begin{subfigure}{4cm}
        \includegraphics[width=4cm]{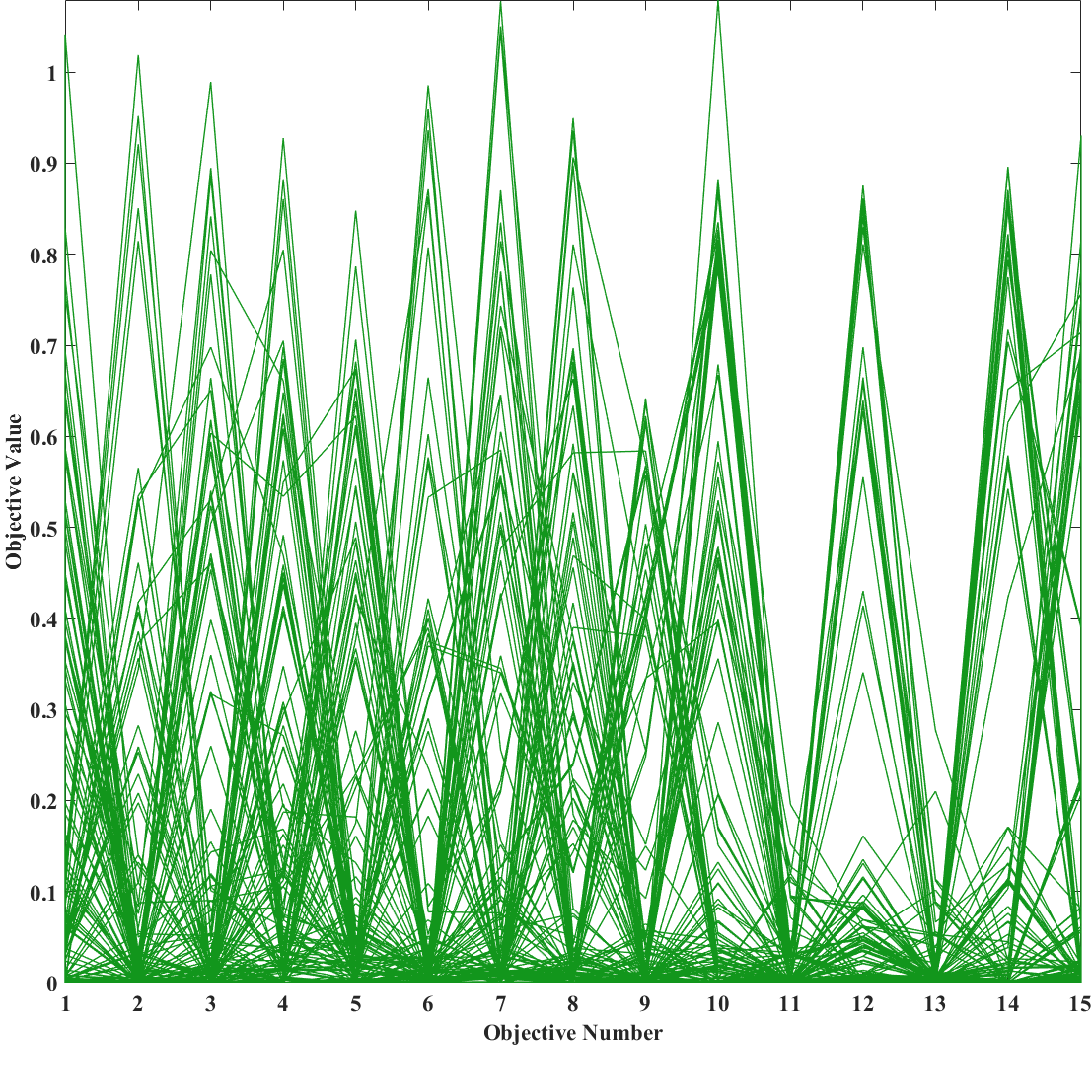}
        \subcaption*{SRA3 for normalized $I_{SDE}$}
    \end{subfigure}
    \begin{subfigure}{4cm}
        \includegraphics[width=4cm]{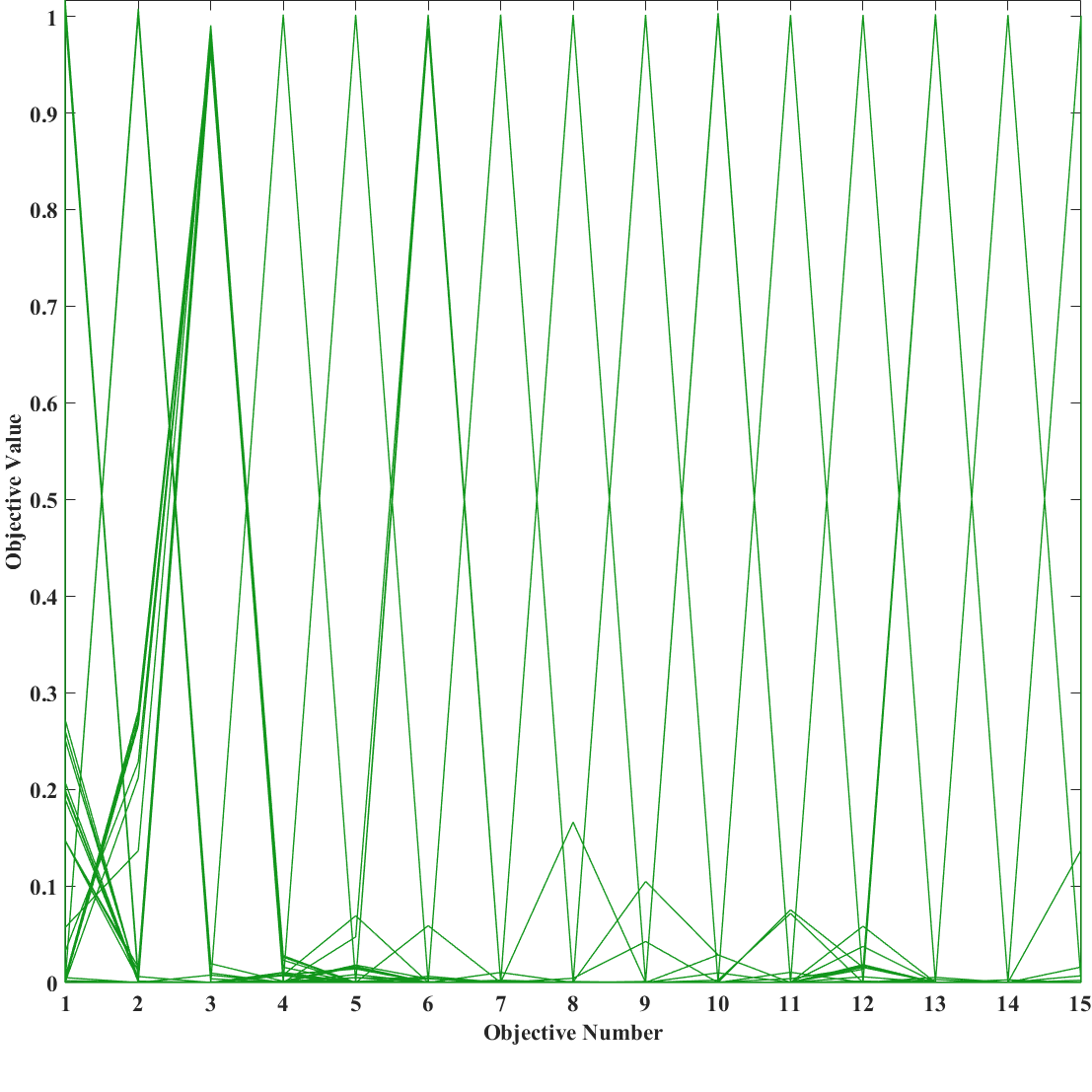}
        \subcaption*{Both normalized SRA3}
    \end{subfigure}
    \caption{Comparison of parallel coordinate plots of four versions of SRA3 on the DTLZ3 problem with 15 objectives.}
\end{figure*}

\begin{figure*}
    \centering
    \begin{subfigure}{4cm}
        \includegraphics[width=4cm]{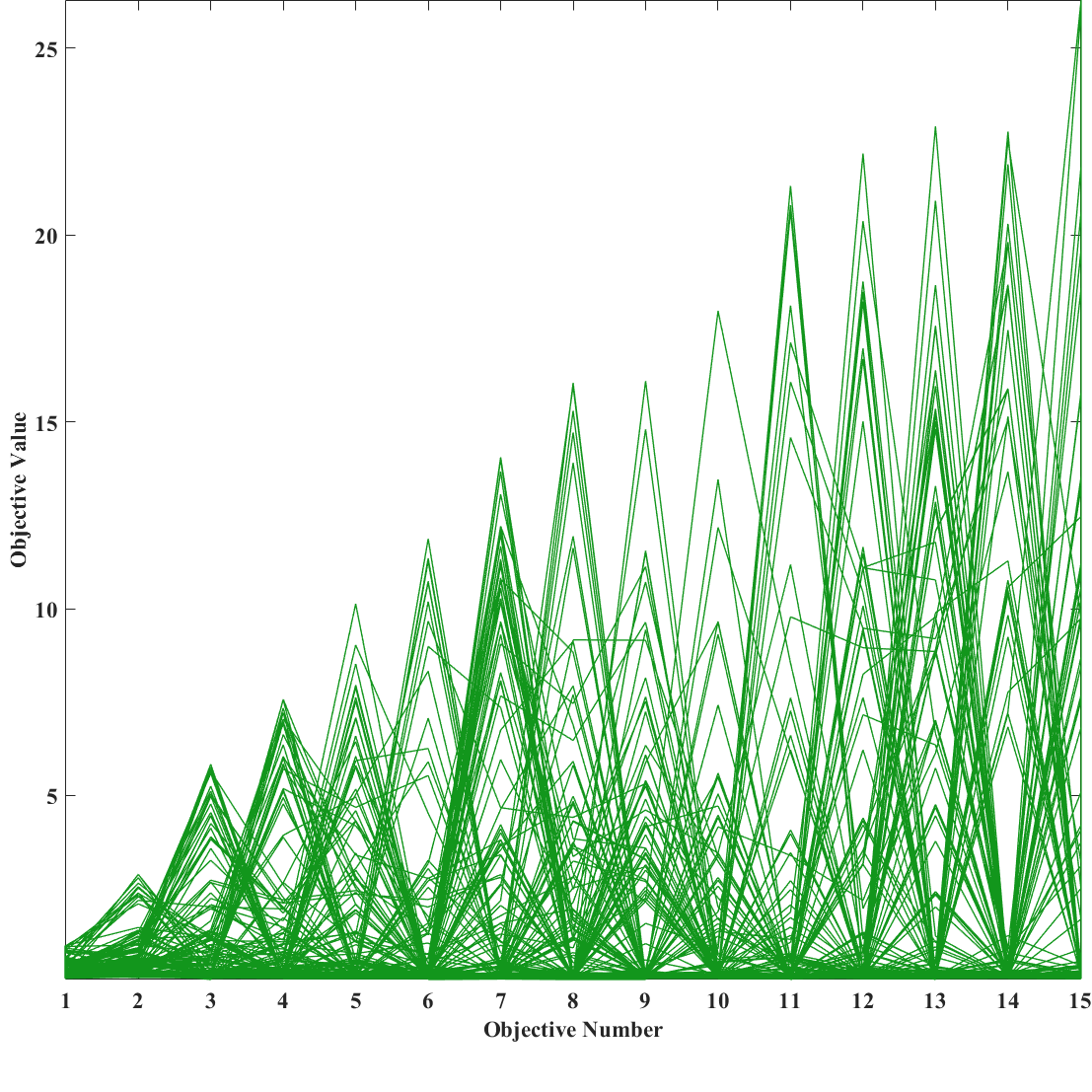}
        \subcaption*{SRA3 without normalized}
    \end{subfigure}
    \begin{subfigure}{4cm}
        \includegraphics[width=4cm]{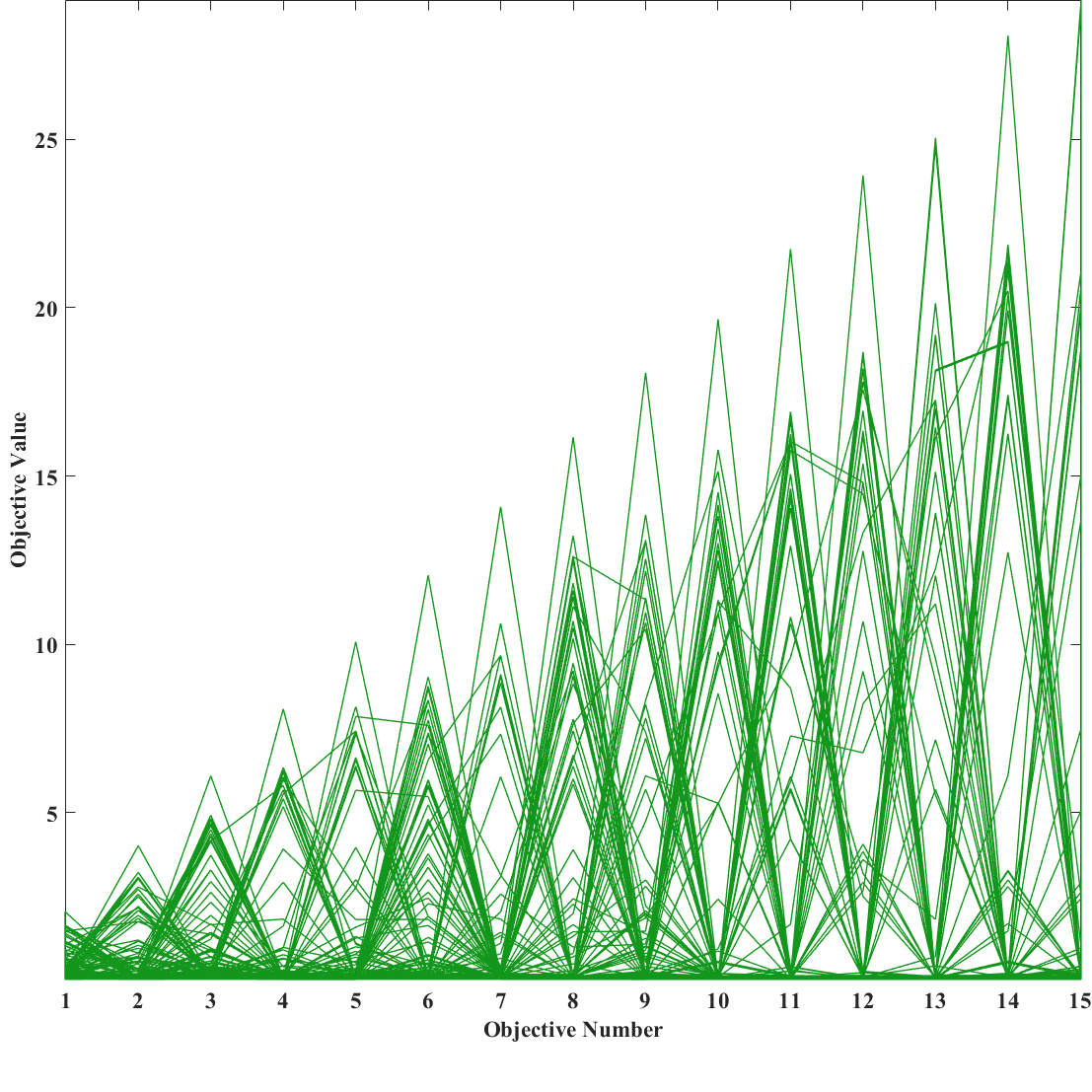}
        \subcaption*{SRA3 for normalized $I_{\epsilon+}$}
    \end{subfigure}
    \begin{subfigure}{4cm}
        \includegraphics[width=4cm]{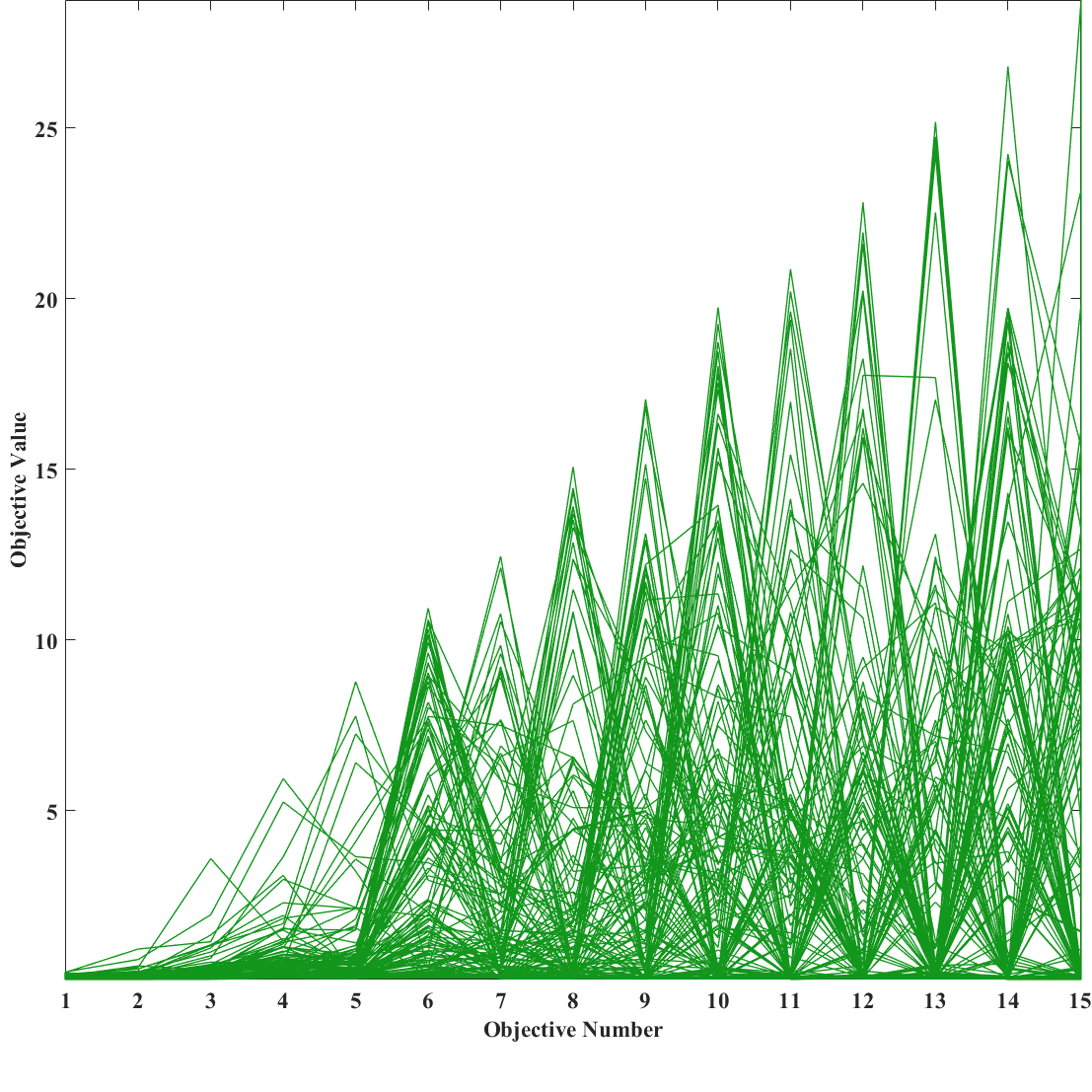}
        \subcaption*{SRA3 for normalized $I_{SDE}$}
    \end{subfigure}
    \begin{subfigure}{4cm}
        \includegraphics[width=4cm]{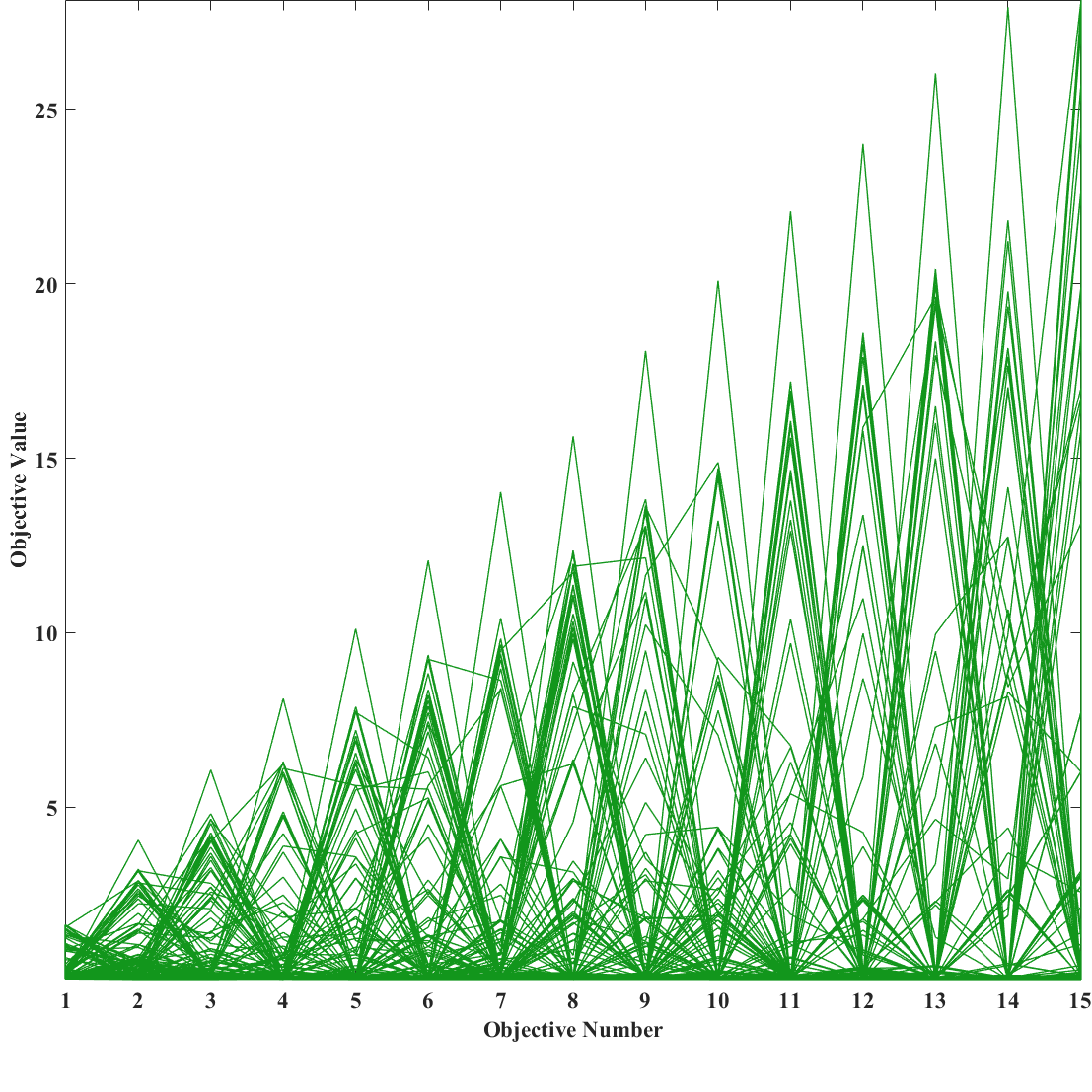}
        \subcaption*{Both normalized SRA3}
    \end{subfigure}
    \caption{Comparison of parallel coordinate plots of four versions of SRA3 on the WFG4 problem with 15 objectives.}
\end{figure*}

\begin{figure*}
    \centering
    \begin{subfigure}{4cm}
        \includegraphics[width=4cm]{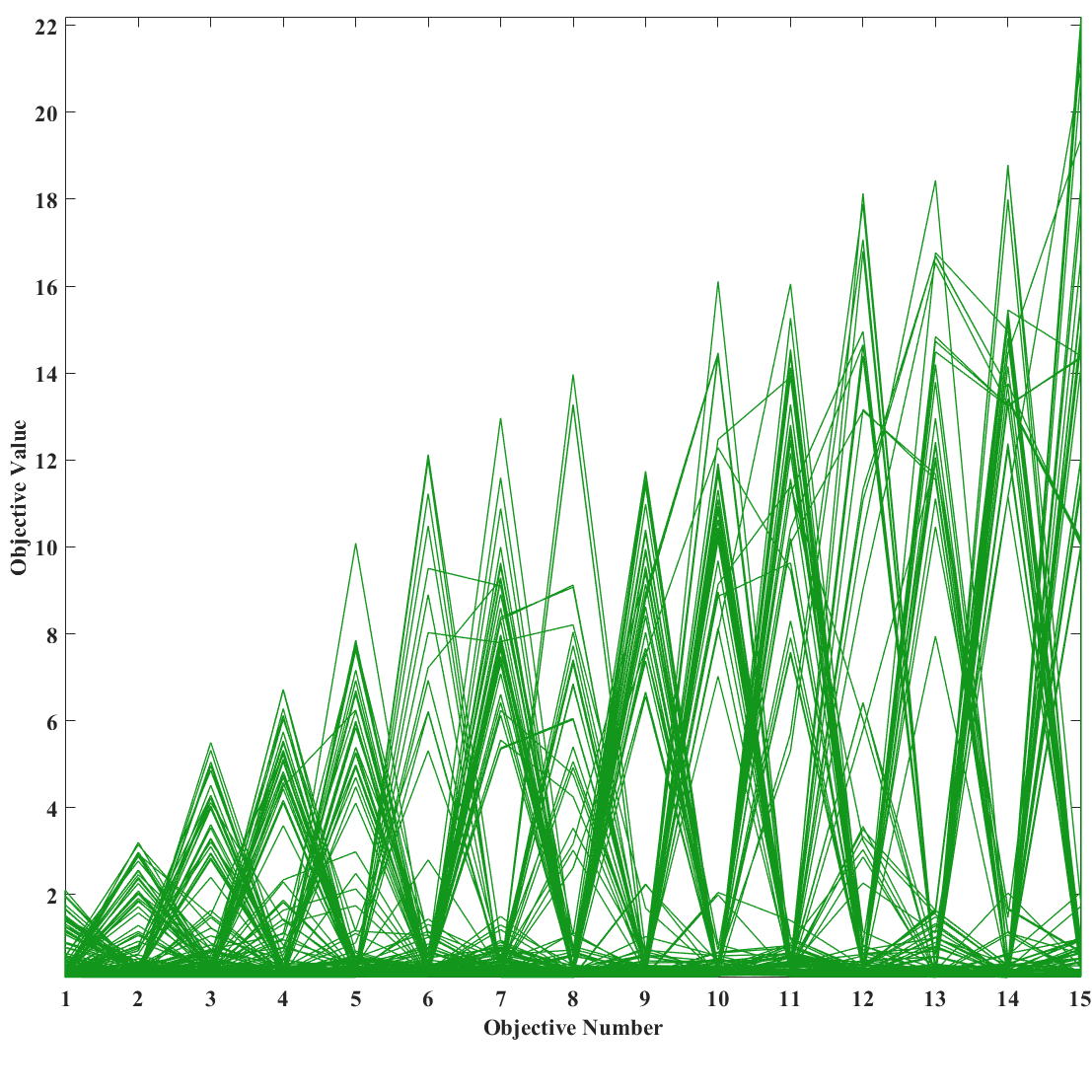}
        \subcaption*{SRA3 without normalized}
    \end{subfigure}
    \begin{subfigure}{4cm}
        \includegraphics[width=4cm]{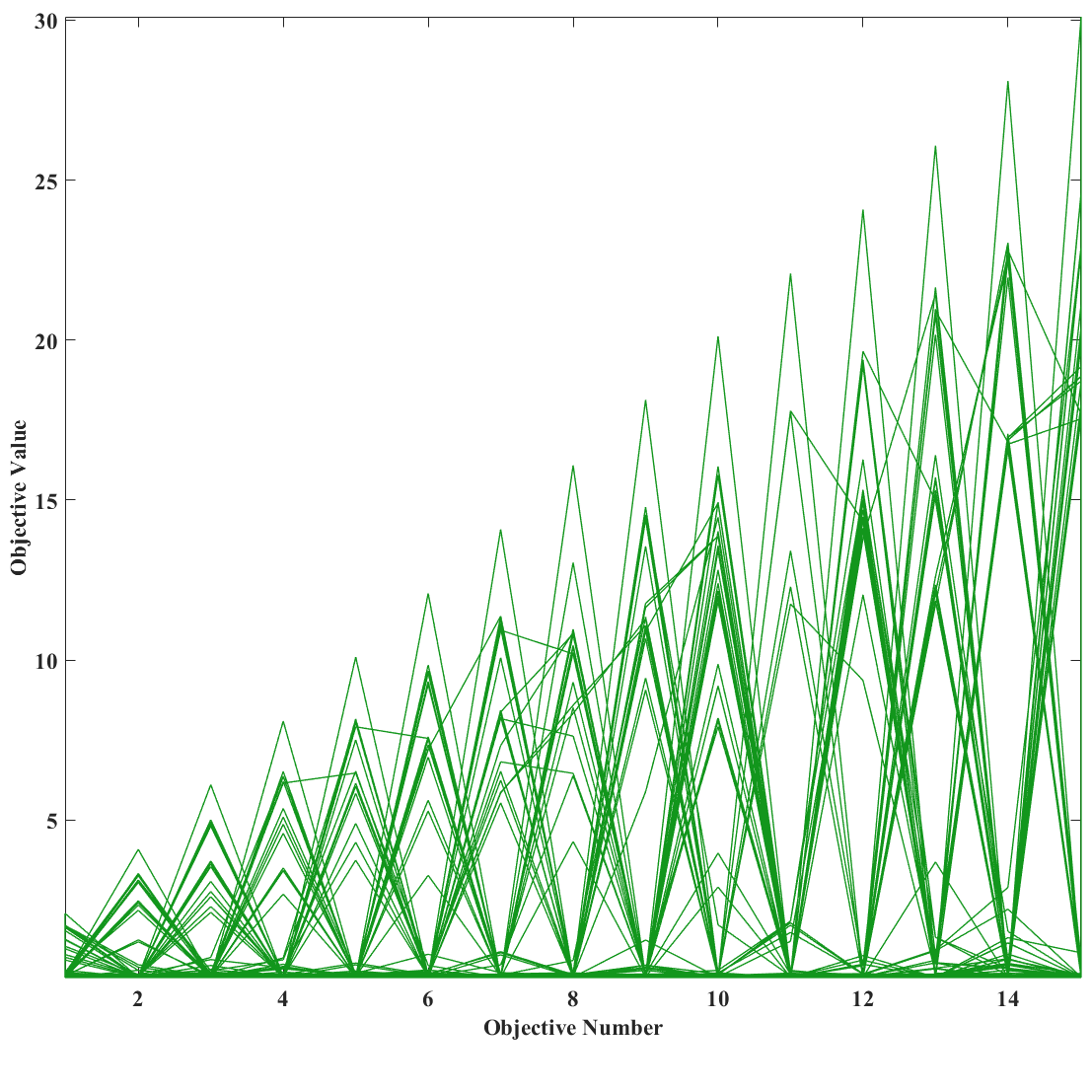}
        \subcaption*{SRA3 for normalized $I_{\epsilon+}$}
    \end{subfigure}
    \begin{subfigure}{4cm}
        \includegraphics[width=4cm]{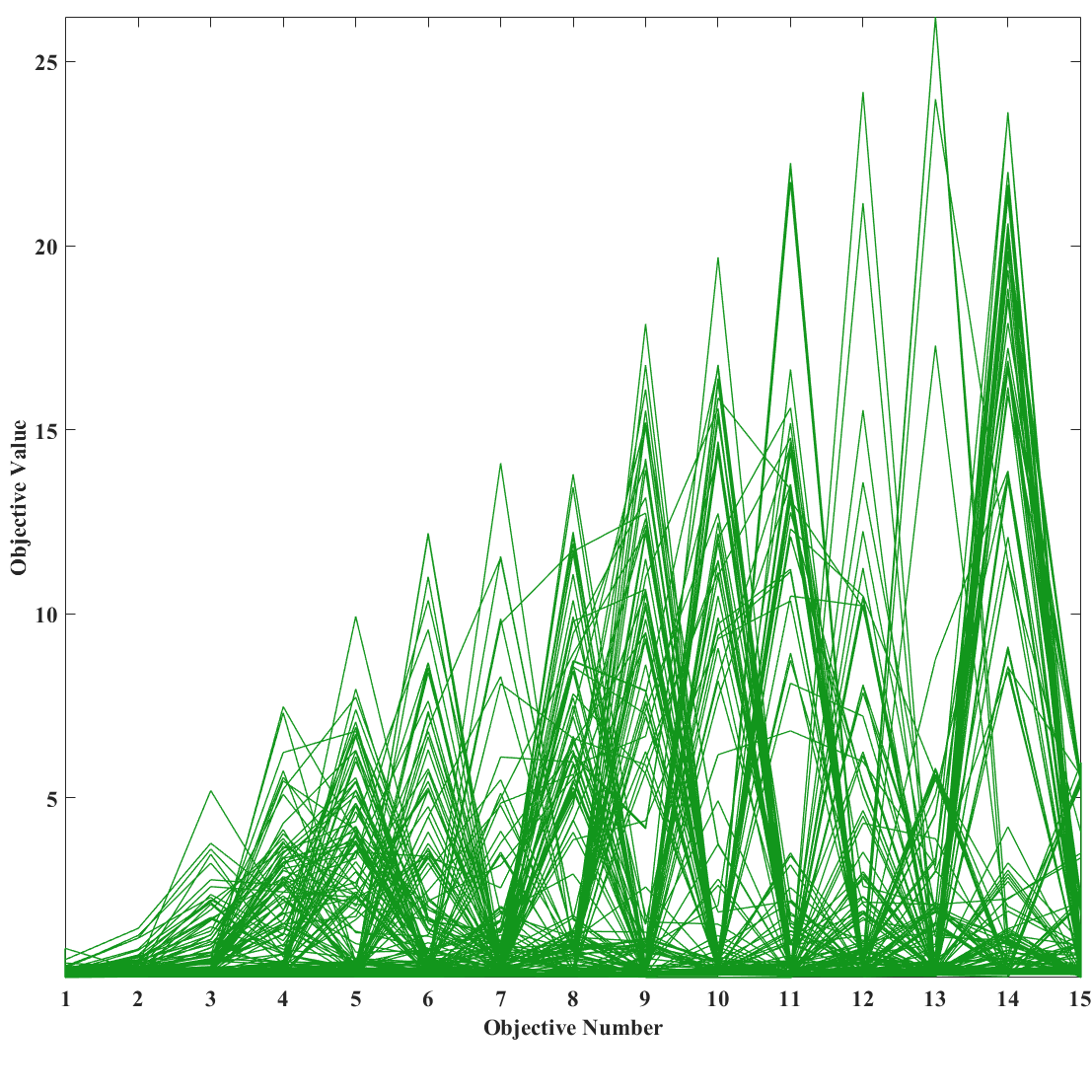}
        \subcaption*{SRA3 for normalized $I_{SDE}$}
    \end{subfigure}
    \begin{subfigure}{4cm}
        \includegraphics[width=4cm]{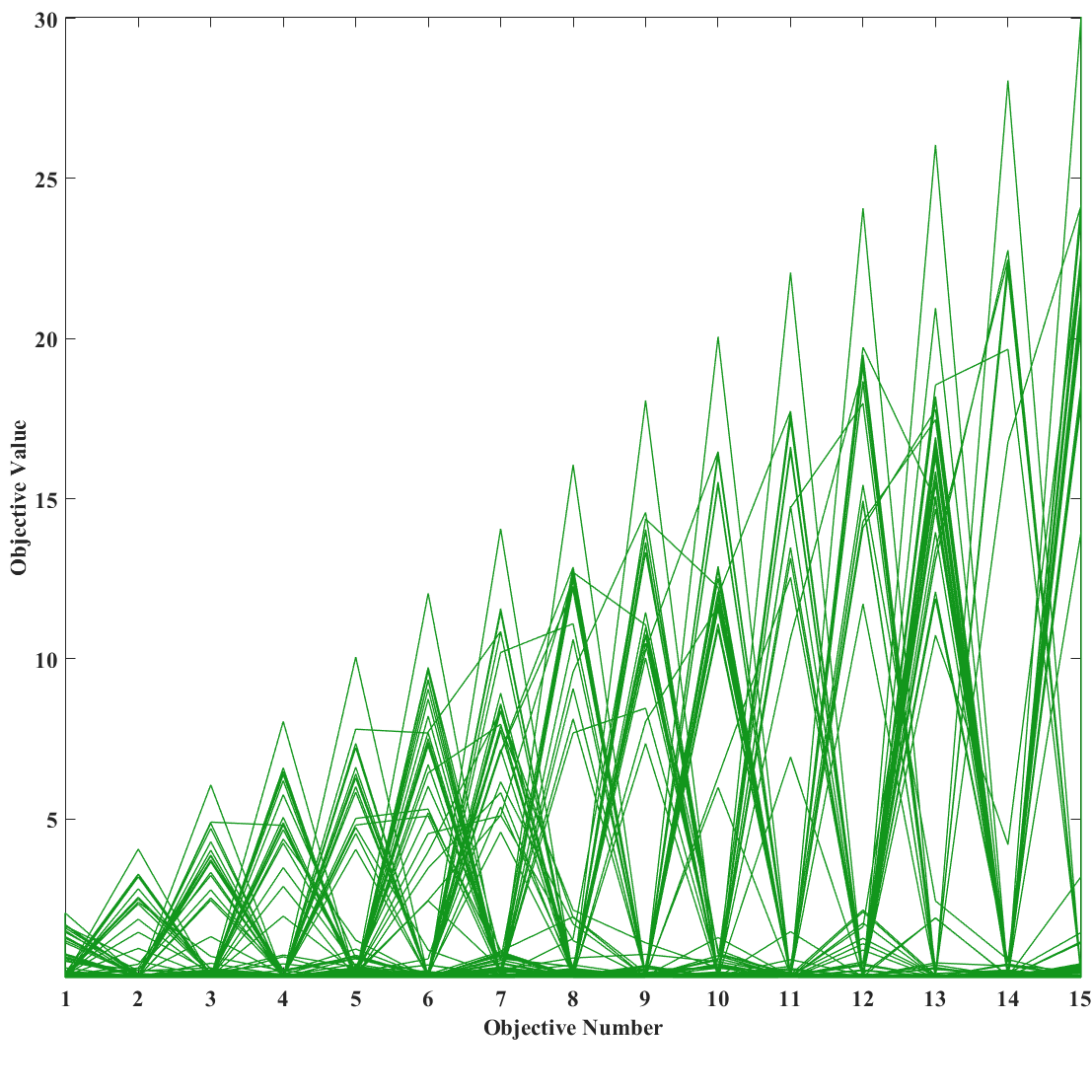}
        \subcaption*{Both normalized SRA3}
    \end{subfigure}
    \caption{Comparison of parallel coordinate plots of four versions of SRA3 on the WFG6 problem with 15 objectives.}
\end{figure*}

Tables 10 and 11 show that, normalizing only the $I_{\epsilon+}$ indicator and normalizing both indicators have similar performance, while normalizing only the $I_{SDE}$ indicator and without normalization also have similar performance. Normalizing only the $I_{\epsilon+}$ indicator and normalizing both indicators perform better on the DTLZ2 and DTLZ4 problems, whereas perform worse on the DTLZ1 and DTLZ3 problems. And as shown in Figures 1 to 4, normalizing only the $I_{\epsilon+}$ indicator and normalizing both indicators lead to the solutions of all problems tend to converge to extreme solutions (including the DTLZ and WFG problems), thus reducing the diversity of solutions, while normalizing only the $I_{SDE}$ indicator does not have this phenomenon. Also, on the DTLZ2 and WFG6 problems, we found that normalizing only the $I_{\epsilon+}$ indicator and normalizing both indicators can find the extreme solutions more easily.

In summary, we argue that the deterioration in the performance of HV and IGD indicators for the DTLZ1 and DTLZ3 problems after normalization is due to the solutions tending to extremes, which is produced by the $I_{\epsilon+}$ indicator (The above phenomenon also exists for SRA and IBEA, but is not shown here due to space constraints). However, for the DTLZ2 and DTLZ4 problems, while normalization also leads to solutions that tend to extremes, extreme solutions are acquired at the same time that are not possible before normalization, resulting in HV and IGD indicators that are better than before normalization.

Then, to verify our conjecture and explore how normalization affects the $I_{\epsilon+}$ indicator, we analyzed the problem on two objectives. Suppose there are 1000 solutions in the space that perform similarly and are normalized, and their distribution are concave, convex, or linear, respectively. Then, the average value of the $I_{\epsilon+}$ indicator was calculated separately for each of the 1000 solutions with similarly performance for concave, convex, or linear distributions, and the results are shown in Figure 5. We also compared the parallel coordinate plots of SRA3 before and after normalization of the $I_{\epsilon+}$ indicator on the DLTZ3 problem with two objectives, and the results are shown in Figure 6 (the size of the CA and DA archives are set to 100).

\begin{figure}
    \centering
    \begin{subfigure}{4cm}
    \includegraphics[width=4cm]{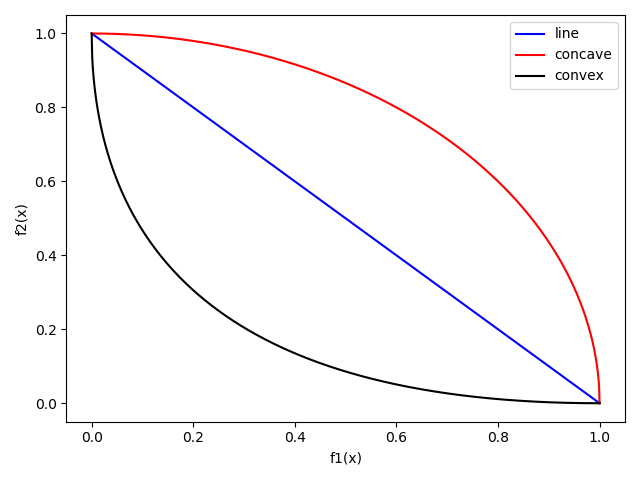}
    \subcaption*{Function Diagram}
    \end{subfigure}
    \begin{subfigure}{4cm}
    \includegraphics[width=4cm]{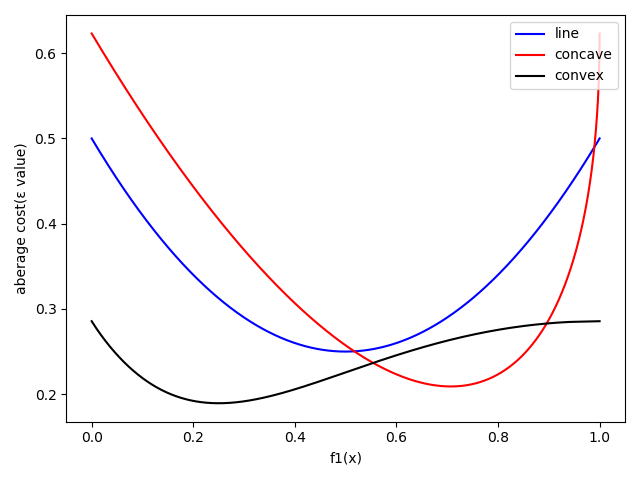}
    \subcaption*{the average value of the $I_{\epsilon+}$}
    \end{subfigure}
    \caption{Distribution of 1000 similarly performing solutions in space after normalization and their average value of the $I_{\epsilon+}$ indicator.}
\end{figure}

\begin{figure}
    \centering
    \begin{subfigure}{4cm}
        \includegraphics[width=4cm]{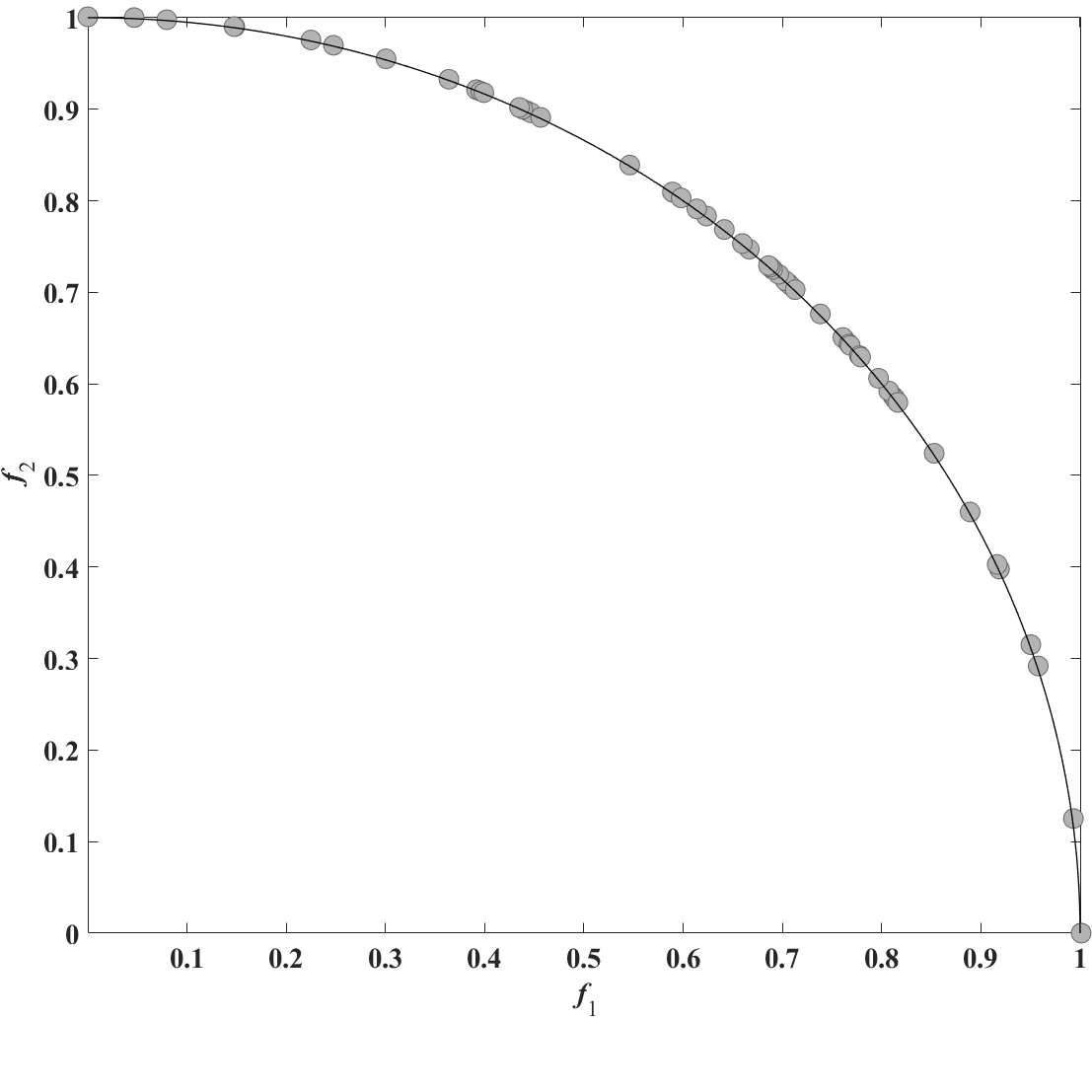}
        \subcaption*{without normalization}
    \end{subfigure}
    \begin{subfigure}{4cm}
        \includegraphics[width=4cm]{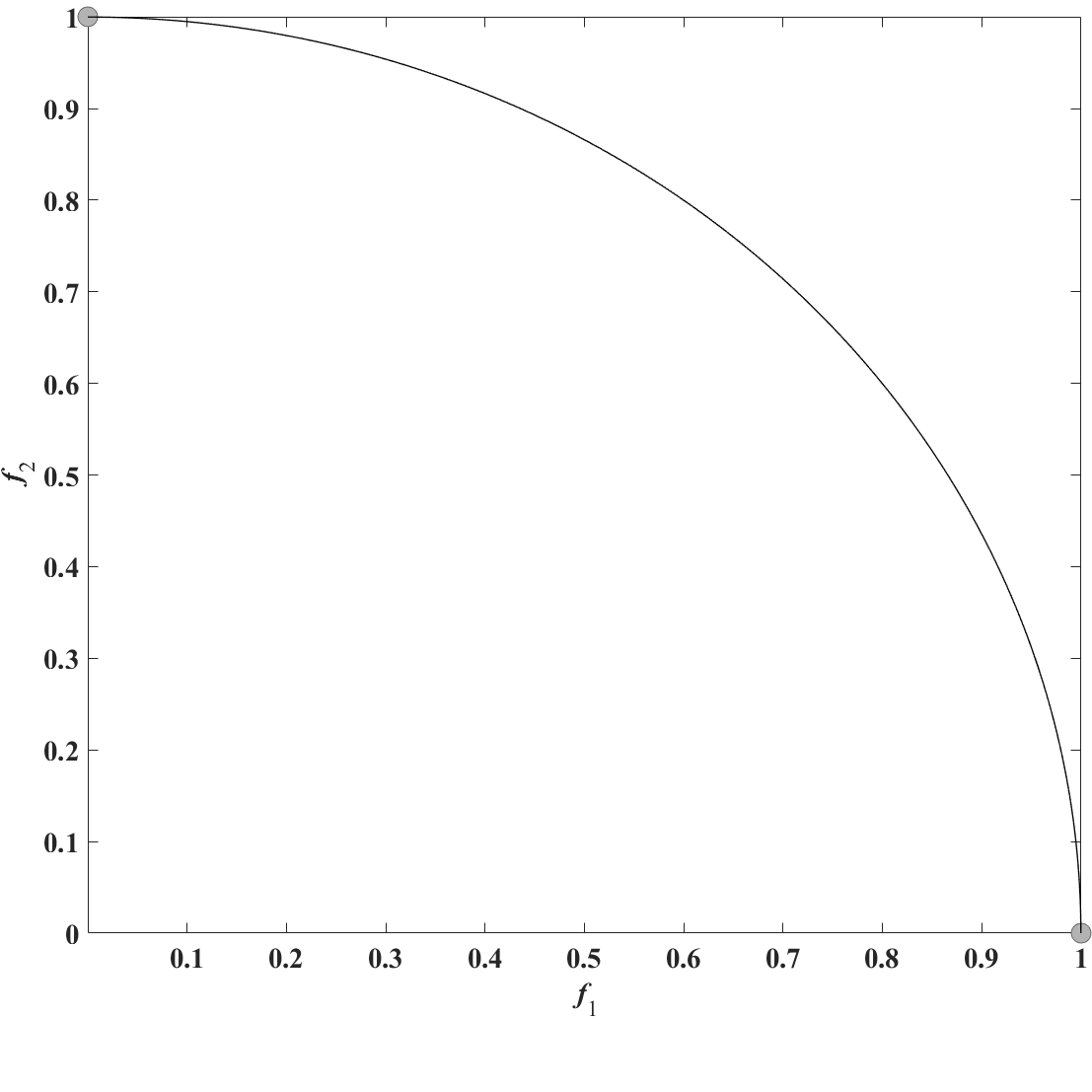}
        \subcaption*{normalized $I_{\epsilon+}$}
    \end{subfigure}
    \caption{Comparison of parallel coordinate plots of SRA3 before and after normalizing the $I_{\epsilon+}$ indicator on the DTLZ3 problem with 2 objectives.}
\end{figure}

Figure 5 shows that after normalizing the $I_{\epsilon+}$ indicator, the average $I_{\epsilon+}$ indicator value of the extreme solutions is greater at any moment in the search process, regardless of whether the distribution of similar-performing solutions is concave, convex, or linear in space. In other words, the normalized $I_{\epsilon+}$ indicator has a preference for the extreme solutions. Also, the parallel coordinate plots in Figure 6 can verify this conclusion from the side. After normalizing the $I_{\epsilon+}$ indicator, SRA3 converges to two extreme solutions (1, 0) and (0, 1) on the two-objective DTLZ3 problem, while a more uniformly distributed solution set can be obtained without normalization.

Then, we continue analyzed the problem on two objectives to explore the bias of the $I_{\epsilon+}$ indicator without normalization. We assumed that there are 1000 solutions in the space that perform similarly, and their distribution are concave, convex, or linear, respectively, where the second objective takes twice the range of values of the first objective. And the average value of the $I_{\epsilon+}$ indicator was calculated separately for each of the 1000 solutions with similarly performance for concave, convex, or linear distributions, and the results are shown in Figure 7. We also analyzed the parallel coordinate plots of SRA3 without normalization of the $I_{\epsilon+}$ indicator on the WFG1 and WFG2 problems with two objectives, and the results are shown in Figure 8 (the archives size of the WFG1 problem is set to 100, and the archives size of the WFG2 problem is set to 30).

\begin{figure}
    \centering
    \begin{subfigure}{4cm}
    \includegraphics[width=4cm]{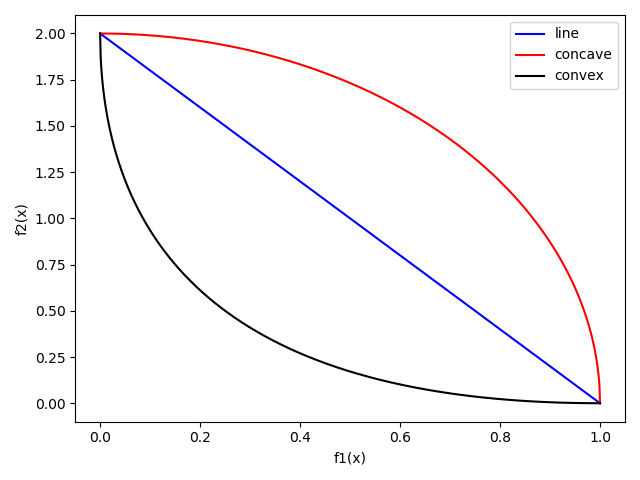}
    \subcaption*{Function Diagram}
    \end{subfigure}
    \begin{subfigure}{4cm}
    \includegraphics[width=4cm]{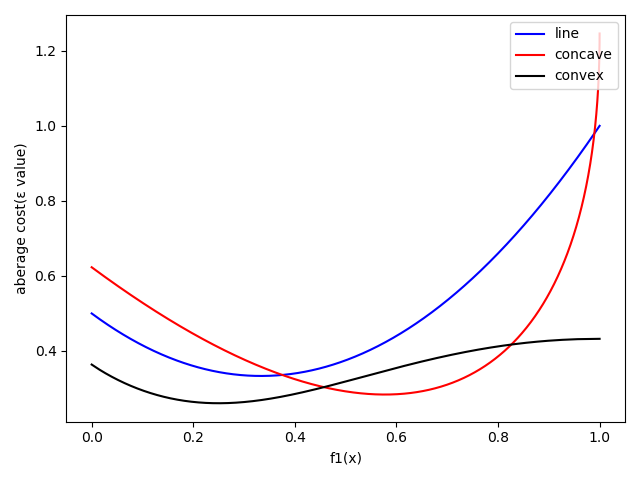}
    \subcaption*{the average value of the $I_{\epsilon+}$}
    \end{subfigure}
    \caption{Distribution of 1000 similarly performing solutions in space without normalization and their average value of the $I_{\epsilon+}$ indicator.}
\end{figure}

\begin{figure}
    \centering
    \begin{subfigure}{4cm}
    \includegraphics[width=4cm]{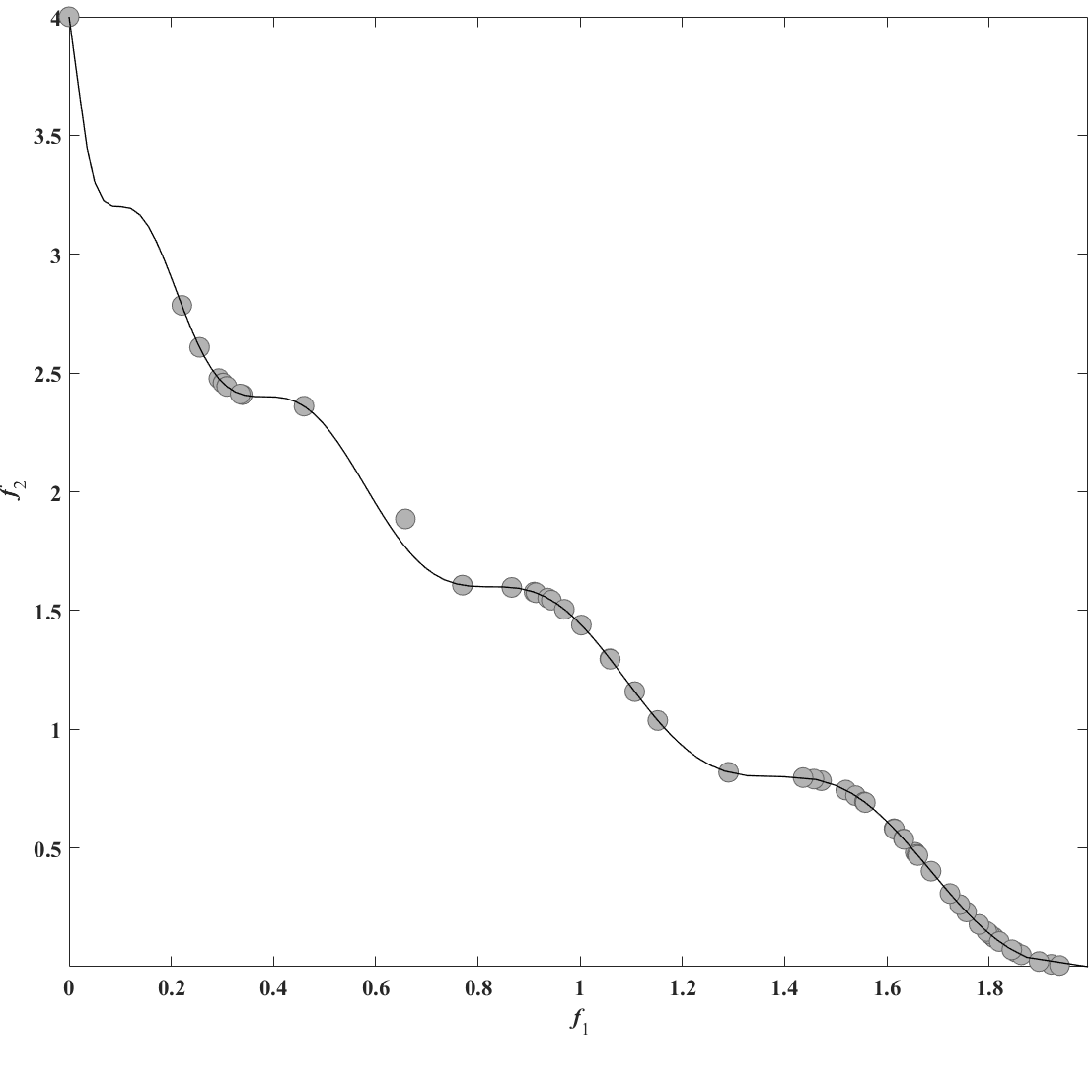}
    \subcaption*{WFG1}
    \end{subfigure}
    \begin{subfigure}{4cm}
    \includegraphics[width=4cm]{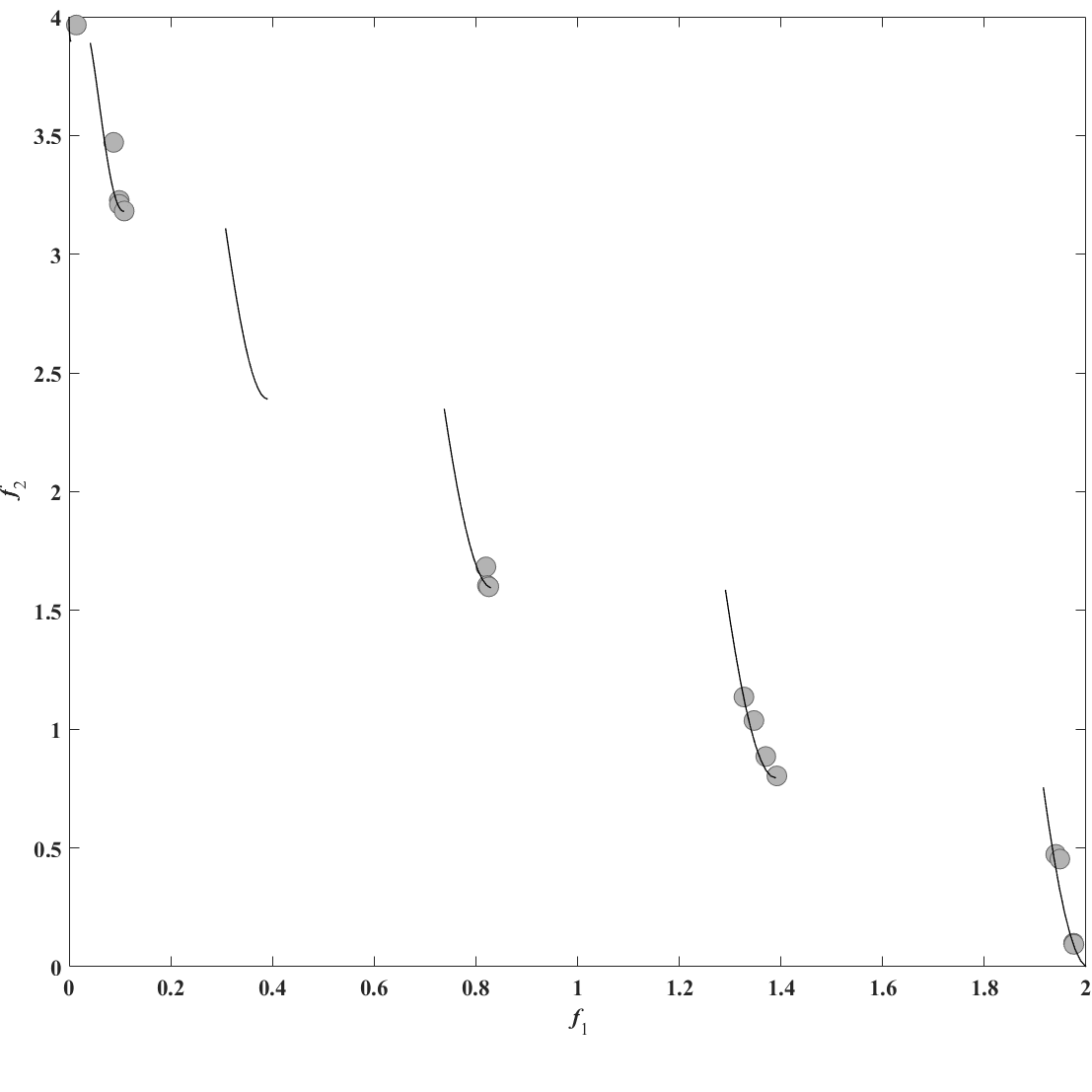}
    \subcaption*{WFG2}
    \end{subfigure}
    \caption{Comparison of parallel coordinate plots of SRA3 without normalized on the WFG1 and WFG2 problems with 2 objectives.}
\end{figure}

Figure 7 shows that when the $I_{\epsilon+}$ indicator is not normalized, it prefers solutions with bigger values on objectives with small value domains, regardless of whether the solution distribution in space is concave, convex, or linear. However, since there is no normalization, different objectives alternate to become objectives with minimum value domains during the search, resulting in a more uniformly distributed solution set. But if each objective in the problem has a different value domain, the solution set will eventually be skewed toward greater values for objectives with smaller value domains, resulting in a biased final conclusion. The parallel coordinate plots in Figure 6 and 8 can be used to validate this conclusion from the side. On the DTLZ3 problem, the solution set is more uniformly distributed without normalization. However, for WFG1 and WFG2, problems with different value domain for each objective, the solution sets are more biased toward higher values on objectives with small value domains.

In summary, when the problem has different value domain for each objective, normalization is required to eliminate the preference of the $I_{\epsilon+}$ indicator for solutions with larger values for a small range of objectives. When each of the problem's objectives has the same value domain, normalizing is recommended if it is difficult to find the extreme solutions, otherwise normalization is not recommended.

\subsubsection{Experimental study of normalized SRA3 on DTLZ and WFG problems}
To further validate the performance of the normalized SRA3 on each problem, SRA3 was compared with the SRA, IBEA, Two$\_$Arch2, NSGA-III, and MOEA/D algorithms on the DTLZ1-4 and WFG1-9 problems with 5, 10, 15 objectives, and evaluated by HV and IGD indicators. To be fair, SRA, IBEA, and Two$\_$Arch2 were normalized (all algorithms performed better after normalization, but are not shown here due to space limitations). Finally, the comparison results are shown in Tables 12 to 15.

\begin{table*}[width=1.94\linewidth,cols=8,pos=h]
\caption{Performance comparison of SRA3 and state-of-the-art algorithms in terms of the average HV values on DTLZ problems. The indicator values are followed by their algorithmic ranking and comparison with the Wilcoxon rank sum test of SRA3 (significance level = 0.05). The best result is highlighted in boldface.}\label{tbl1}
\begin{tabular*}{\tblwidth}{@{} CCCCCCCC@{}}
\toprule
Problem & $m$ & SRA3$_{norm}$ & SRA$_{norm}$ & IBEA$_{norm}$ & Two$\_$Arch2$_{norm}$ & NSGA-III & MOEA/D \\
\midrule
DTLZ1 & 5 & 8.58$e-$01(5) & 9.32$e-$01(4)+ & 7.47$e-$01(6)$-$ & 9.76$e-$01(3)+ & \textbf{9.80$e-$01(1)+} & 9.79$e-$01(2)+ \\
DTLZ1 & 10 & 9.72$e-$01(4) & \textbf{9.99$e-$01(1)+} & 9.34$e-$01(5)$-$ & 9.95$e-$01(2)+ & 9.33$e-$01(6)$-$ & 9.88$e-$01(3)+\\
DTLZ1 & 15 & 8.74$e-$01(5) & \textbf{9.97$e-$01(1)+} & 7.82$e-$01(6)$-$ & 9.86$e-$01(2)+ & 9.60$e-$01(3)+ & 9.55$e-$01(4)+\\
%\specialrule{0em}{0pt}{0.2pt}
\hdashline \specialrule{0em}{0pt}{0.02cm}
%\specialrule{0em}{0pt}{0.2pt}
DTLZ2 & 5 & 8.05$e-$01(5) & 8.08$e-$01(4)+ & 8.09$e-$01(3)+ & 7.67$e-$01(6)$-$ & 8.12$e-$01(2)+ & \textbf{8.12$e-$01(1)+}\\
DTLZ2 & 10 & 9.73$e-$01(2) & 9.50$e-$01(5)$-$ & \textbf{9.74$e-$01(1)+} & 7.40$e-$01(6)$-$ & 9.60$e-$01(4)$-$ & 9.70$e-$01(3)$-$\\
DTLZ2 & 15 & 9.85$e-$01(2) & 9.33$e-$01(5)$-$ & 9.85$e-$01(3)= & 6.10$e-$01(6)$-$ & 9.80$e-$01(4)= & \textbf{9.90$e-$01(1)+}\\
%\specialrule{0em}{0pt}{0.2pt}
\hdashline 
\specialrule{0em}{0pt}{0.02cm}
%\specialrule{0em}{0pt}{0.2pt}
DTLZ3 & 5 & 3.80$e-$01(5) & 7.60$e-$01(2)+ & 3.77$e-$01(6)$-$ & 7.47$e-$01(3)+ & 7.03$e-$01(4)+ & \textbf{7.84$e-$01(1)+}\\
DTLZ3 & 10 & 6.15$e-$01(2) & \textbf{9.22$e-$01(1)+} & 6.11$e-$01(3)$-$ & 6.10$e-$01(4)= & 2.36$e-$02(6)$-$ & 5.31$e-$01(5)=\\
DTLZ3 & 15 & 7.07$e-$01(2) & \textbf{9.12$e-$01(1)+} & 5.51$e-$01(3)$-$ & 0.00$e+$00(6)$-$ & 1.32$e-$01(5)$-$ & 5.38$e-$01(4)=\\
%\specialrule{0em}{0pt}{0.2pt}
\hdashline \specialrule{0em}{0pt}{0.02cm}
%\specialrule{0em}{0pt}{0.2pt}
DTLZ4 & 5 & 8.06$e-$01(4) & \textbf{8.12$e-$01(1)+} & 8.09$e-$01(2)+ & 7.54$e-$01(5)$-$ & 8.06$e-$01(3)+ & 6.63$e-$01(6)$-$\\
DTLZ4 & 10 & 9.73$e-$01(2) & 9.62$e-$01(4)$-$ & \textbf{9.74$e-$01(1)+} & 7.18$e-$01(6)$-$ & 9.67$e-$01(3)$-$ & 8.99$e-$01(5)$-$\\
DTLZ4 & 15 & \textbf{9.87$e-$01(1)} & 9.62$e-$01(4)$-$ & {9.87$e-$01(2)=} & 6.33$e-$01(6)$-$ & 9.81$e-$01(3)= & 9.13$e-$01(5)$-$\\
\hline
    \multicolumn{2}{c}{Win/Tie/Loss} & - & 8/0/4 & 4/2/6 & 4/1/7 & 5/2/5 & 6/2/4 \\
\bottomrule
\end{tabular*}
\end{table*}

\begin{table*}[width=1.94\linewidth,cols=8,pos=h]
\caption{Performance comparison of SRA3 and state-of-the-art algorithms in terms of the average IGD values on DTLZ problems. The indicator values are followed by their algorithmic ranking and comparison with the Wilcoxon rank sum test of SRA3 (significance level = 0.05). The best result is highlighted in boldface.}\label{tbl1}
\begin{tabular*}{\tblwidth}{@{} CCCCCCCC@{} }
\toprule
Problem & $m$ & SRA3$_{norm}$ & SRA$_{norm}$ & IBEA$_{norm}$ & Two$\_$Arch2$_{norm}$ & NSGA-III & MOEA/D \\
\midrule
DTLZ1 & 5 & 2.82$e-$01(5) & 1.40$e-$01(4)+ & 3.46$e-$01(6)$-$ & \textbf{9.64$e-$02(1)+} & 9.93$e-$02(2)+ & 9.96$e-$02(3)+\\
DTLZ1 & 10 & 3.87$e-$01(5) & 2.07$e-$01(3)+ & 4.30$e-$01(6)$-$ & 1.87$e-$01(2)+ & 2.98$e-$01(4)+ & \textbf{1.75$e-$01(1)+}\\
DTLZ1 & 15 & 5.55$e-$01(5) & 3.61$e-$01(4)+ & 6.34$e-$01(6)$-$ & 2.76$e-$01(2)+ & 3.58$e-$01(3)+ & \textbf{2.36$e-$01(1)+}\\
%\specialrule{0em}{0pt}{0.2pt}
\hdashline \specialrule{0em}{0pt}{0.02cm}
%\specialrule{0em}{0pt}{0.2pt}
DTLZ2 & 5 & 1.73$e-$01(5) & 1.69$e-$01(4)+ & 1.76$e-$01(6)$-$ & \textbf{1.52$e-$01(1)+} & 1.58$e-$01(3)+ & 1.58$e-$01(2)+\\
DTLZ2 & 10 & 3.85$e-$01(4) & 3.68$e-$01(2)+ & 3.88$e-$01(5)$-$ & 3.95$e-$01(6)$-$ & 3.80$e-$01(3)+ & \textbf{3.56$e-$01(1)+}\\
DTLZ2 & 15 & 5.52$e-$01(3) & \textbf{5.19$e-$01(1)+} & 5.76$e-$01(5)$-$ & 5.63$e-$01(4)$-$ & 5.82$e-$01(6)= & 5.45$e-$01(2)+\\
%\specialrule{0em}{0pt}{0.2pt}
\hdashline \specialrule{0em}{0pt}{0.02cm}
%\specialrule{0em}{0pt}{0.2pt}
DTLZ3 & 5 & 5.35$e-$01(5) & 2.35$e-$01(3)+ & 5.58$e-$01(6)$-$ & 1.89$e-$01(2)+ & 2.65$e-$01(4)+ & \textbf{1.87$e-$01(1)+}\\
DTLZ3 & 10 & 6.49$e-$01(3) & \textbf{4.46$e-$01(1)+} & 6.61$e-$01(4)$-$ & 6.22$e-$01(2)+ & 3.69$e+$00(6)$-$ & 7.06$e-$01(5)=\\
DTLZ3 & 15 & 7.51$e-$01(2) & \textbf{6.32$e-$01(1)+} & 8.47$e-$01(4)$-$ & 2.28$e+$01(6)$-$ & 2.57$e+$00(5)$-$ & 8.40$e-$01(3)=\\
%\specialrule{0em}{0pt}{0.2pt}
\hdashline \specialrule{0em}{0pt}{0.02cm}
%\specialrule{0em}{0pt}{0.2pt}
DTLZ4 & 5 & 1.71$e-$01(4) & 1.71$e-$01(3)= & 1.72$e-$01(5)$-$ & \textbf{1.53$e-$01(1)+} & 1.69$e-$01(2)+ & 4.17$e-$01(6)$-$\\
DTLZ4 & 10 & 3.83$e-$01(3) & \textbf{3.72$e-$01(1)+} & 3.91$e-$01(5)$-$ & 3.87$e-$01(4)$-$ & 3.75$e-$01(2)+ & 5.49$e-$01(6)$-$\\
DTLZ4 & 15 & 5.53$e-$01(3) & \textbf{5.39$e-$01(1)+} & 5.75$e-$01(4)$-$ & 5.51$e-$01(2)= & 5.88$e-$01(5)$-$ & 7.22$e-$01(6)$-$\\
\hline
    \multicolumn{2}{c}{Win/Tie/Loss} & - & 11/0/1 & 0/0/12 & 7/1/4 & 8/1/3 & 7/2/3 \\
\bottomrule
\end{tabular*}
\end{table*}

\begin{table*}[width=1.94\linewidth,cols=8,pos=h!]
\caption{Performance comparison of SRA3 and state-of-the-art algorithms in terms of the average HV values on WFG problems. The indicator values are followed by their algorithmic ranking and comparison with the Wilcoxon rank sum test of SRA3 (significance level = 0.05). The best result is highlighted in boldface.}\label{tbl1}
\begin{tabular*}{\tblwidth}{@{} CCCCCCCC@{} }
\toprule
Problem & $m$ & SRA3$_{norm}$ & SRA$_{norm}$ & IBEA$_{norm}$ & Two$\_$Arch2$_{norm}$ & NSGA-III & MOEA/D \\
\midrule
WFG1 & 5 & 9.90$e-$01(2) & 9.77$e-$01(5)= & 9.88$e-$01(3)$-$ & \textbf{9.96$e-$01(1)+} & 9.86$e-$01(4)$-$ & 9.44$e-$01(6)$-$\\
WFG1 & 10 & 9.94$e-$01(3) & 9.83$e-$01(4)= & 9.94$e-$01(2)= & \textbf{9.97$e-$01(1)+} & 9.90$e-$01(6)$-$ & 9.19$e-$01(5)$-$\\
WFG1 & 15 & 9.92$e-$01(3) & 9.87$e-$01(5)= & 9.90$e-$01(4)= & 9.98$e-$01(2)+ & \textbf{1.00$e+$00(1)+} & 8.52$e-$01(6)$-$\\
%\specialrule{0em}{0pt}{0.2pt}
\hdashline \specialrule{0em}{0pt}{0.02cm}
%\specialrule{0em}{0pt}{0.2pt}
WFG2 & 5 & 9.88$e-$01(3) & 9.82$e-$01(5)$-$ & 9.82$e-$01(4)$-$ & \textbf{9.95$e-$01(1)+} & 9.93$e-$01(2)+ & 9.56$e-$01(6)$-$\\
WFG2 & 10 & 9.93$e-$01(3) & 9.91$e-$01(4)$-$ & 9.85$e-$01(5)$-$ & \textbf{9.95$e-$01(1)+} & 9.93$e-$01(2)= & 9.26$e-$01(6)$-$\\
WFG2 & 15 & 9.91$e-$01(3) & 9.89$e-$01(4)$-$ & 9.79$e-$01(5)$-$ & \textbf{9.97$e-$01(1)+} & 9.94$e-$01(2)+ & 9.18$e-$01(6)$-$\\
%\specialrule{0em}{0pt}{0.2pt}
\hdashline \specialrule{0em}{0pt}{0.02cm}
%\specialrule{0em}{0pt}{0.2pt}
WFG3 & 5 & 2.36$e-$01(2) & 1.28$e-$01(5)$-$ & \textbf{2.63$e-$01(1)+} & 1.99$e-$01(3)$-$ & 1.57$e-$01(4)$-$ & 3.19$e-$02(6)$-$\\
WFG3 & 10 & 0.00$e+$00(-) & 0.00$e+$00(-)= & \textbf{5.94$e-$03(1)+} & 4.12$e-$04(2)= & 0.00$e+$00(-)= & 0.00$e+$00(-)$=$\\
WFG3 & 15 & 0.00$e+$00(-) & 0.00$e+$00(-)= & 0.00$e+$00(-)= & 0.00$e+$00(-)= & 0.00$e+$00(-)= & 0.00$e+$00(-)$=$\\
%\specialrule{0em}{0pt}{0.2pt}
\hdashline \specialrule{0em}{0pt}{0.02cm}
%\specialrule{0em}{0pt}{0.2pt}
WFG4 & 5 & 7.83$e-$01(3) & 7.76$e-$01(4)$-$ & \textbf{8.02$e-$01(1)+} & 7.53$e-$01(5)$-$ & 7.97$e-$01(2)+ & 7.08$e-$01(6)$-$ \\
WFG4 & 10 & 9.23$e-$01(3) & 8.64$e-$01(4)$-$ & \textbf{9.57$e-$01(1)+} & 8.09$e-$01(5)$-$ & 9.28$e-$01(2)+ & 4.09$e-$01(6)$-$ \\
WFG4 & 15 & 8.83$e-$01(3) & 8.32$e-$01(4)$-$ & \textbf{9.82$e-$01(1)+} & 8.08$e-$01(5)$-$ & 9.77$e-$01(2)+ & 3.09$e-$01(6)$-$ \\
%\specialrule{0em}{0pt}{0.2pt}
\hdashline \specialrule{0em}{0pt}{0.02cm}
%\specialrule{0em}{0pt}{0.2pt}
WFG5 & 5 & 7.42$e-$01(3) & 7.39$e-$01(4)$-$ & 7.56$e-$01(2)+ & 7.10$e-$01(5)$-$ & \textbf{7.59$e-$01(1)+} & 6.76$e-$01(6)$-$ \\
WFG5 & 10 & 8.84$e-$01(3) & 8.15$e-$01(4)$-$ & \textbf{9.01$e-$01(1)+} & 7.64$e-$01(5)$-$ & 8.90$e-$01(2)+ & 4.86$e-$01(6)$-$ \\
WFG5 & 15 & 8.62$e-$01(3) & 7.49$e-$01(4)$-$ & 9.12$e-$01(2)+ & 6.87$e-$01(5)$-$ & \textbf{9.13$e-$01(1)+} & 3.41$e-$01(6)$-$ \\
%\specialrule{0em}{0pt}{0.2pt}
\hdashline \specialrule{0em}{0pt}{0.02cm}
%\specialrule{0em}{0pt}{0.2pt}
WFG6 & 5 & 7.31$e-$01(3) & 7.23$e-$01(4)$-$ & \textbf{7.44$e-$01(1)+} & 6.99$e-$01(5)$-$ & 7.38$e-$01(2)= & 5.94$e-$01(6)$-$ \\
WFG6 & 10 & 8.65$e-$01(3) & 7.96$e-$01(4)$-$ & \textbf{8.87$e-$01(1)+} & 7.39$e-$01(5)$-$ & 8.66$e-$01(2)= & 2.55$e-$01(6)$-$ \\
WFG6 & 15 & 8.59$e-$01(3) & 7.27$e-$01(4)$-$ & \textbf{8.90$e-$01(1)+} & 7.26$e-$01(5)$-$ & 8.89$e-$01(2)+ & 1.30$e-$01(6)$-$ \\
%\specialrule{0em}{0pt}{0.2pt}
\hdashline \specialrule{0em}{0pt}{0.02cm}
%\specialrule{0em}{0pt}{0.2pt}
WFG7 & 5 & 7.96$e-$01(3) & 7.90$e-$01(4)$-$ & \textbf{8.08$e-$01(1)+} & 7.70$e-$01(5)$-$ & 8.03$e-$01(2)+ & 6.55$e-$01(6)$-$ \\
WFG7 & 10 & 9.53$e-$01(2) & 8.87$e-$01(4)$-$ & \textbf{9.69$e-$01(1)+} & 8.35$e-$01(5)$-$ & 9.40$e-$01(3)$-$ & 3.43$e-$01(6)$-$ \\
WFG7 & 15 & 9.72$e-$01(3) & 8.50$e-$01(4)$-$ & \textbf{9.87$e-$01(1)+} & 7.90$e-$01(5)$-$ & 9.81$e-$01(2)+ & 1.64$e-$01(6)$-$ \\
%\specialrule{0em}{0pt}{0.2pt}
\hdashline \specialrule{0em}{0pt}{0.02cm}
%\specialrule{0em}{0pt}{0.2pt}
WFG8 & 5 & 6.77$e-$01(3) & 6.57$e-$01(4)$-$ & \textbf{7.01$e-$01(1)+} & 6.36$e-$01(5)$-$ & 6.87$e-$01(2)+ & 5.47$e-$01(6)$-$ \\
WFG8 & 10 & 8.64$e-$01(2) & 7.38$e-$01(4)$-$ & \textbf{9.16$e-$01(1)+} & 6.11$e-$01(5)$-$ & 8.47$e-$01(3)$-$ & 5.40$e-$02(6)$-$ \\
WFG8 & 15 & 8.33$e-$01(3) & 6.07$e-$01(4)$-$ & 9.07$e-$01(2)+ & 5.62$e-$01(5)$-$ & \textbf{9.13$e-$01(1)+} & 7.90$e-$02(6)$-$ \\
%\specialrule{0em}{0pt}{0.2pt}
\hdashline \specialrule{0em}{0pt}{0.02cm}
%\specialrule{0em}{0pt}{0.2pt}
WFG9 & 5 & 7.34$e-$01(4) & \textbf{7.60$e-$01(1)+} & 7.59$e-$01(2)+ & 7.20$e-$01(5)$-$ & 7.47$e-$01(3)+ & 6.09$e-$01(6)$-$ \\
WFG9 & 10 & 8.25$e-$01(4) & 8.34$e-$01(3)= & \textbf{8.78$e-$01(1)+} & 6.86$e-$01(5)$-$ & 8.35$e-$01(2)+ & 2.88$e-$01(6)$-$ \\
WFG9 & 15 & 7.60$e-$01(3) & 7.55$e-$01(4)$-$ & 8.61$e-$01(2)+ & 6.14$e-$01(5)$-$ & \textbf{8.70$e-$01(1)+} & 2.84$e-$01(6)$-$ \\
\hline
    \multicolumn{2}{c}{Win/Tie/Loss} & - & 1/6/20 & 20/3/4 & 6/2/19 & 17/5/5 & 0/2/25 \\
\bottomrule
\end{tabular*}
\end{table*}

\begin{table*}[width=1.94\linewidth,cols=8,pos=h!]
\caption{Performance comparison of SRA3 and state-of-the-art algorithms in terms of the average IGD values on WFG problems. The indicator values are followed by their algorithmic ranking and comparison with the Wilcoxon rank sum test of SRA3 (significance level = 0.05). The best result is highlighted in boldface.}\label{tbl1}
\begin{tabular*}{\tblwidth}{@{} CCCCCCCC@{} }
\toprule
Problem & $m$ & SRA3$_{norm}$ & SRA$_{norm}$ & IBEA$_{norm}$ & Two$\_$Arch2$_{norm}$ & NSGA-III & MOEA/D \\
\midrule
WFG1 & 5 & 5.77$e-$02(2) & 6.02$e-$02(5)$-$ & 6.02$e-$02(4)$-$ & \textbf{5.27$e-$02(1)+} & 5.92$e-$02(3)= & 9.73$e-$02(6)$-$ \\
WFG1 & 10 & 8.49$e-$02(3) & 8.69$e-$02(4)$-$ & 7.98$e-$02(2)+ & \textbf{7.74$e-$02(1)+} & 1.04$e-$01(5)$-$ & 1.26$e-$01(6)$-$ \\
WFG1 & 15 & 1.05$e-$01(4) & 1.06$e-$01(5)= & 9.24$e-$02(2)+ & \textbf{8.35$e-$02(1)+} & 9.46$e-$02(3)+ & 1.37$e-$01(6)$-$ \\
%\specialrule{0em}{0pt}{0.2pt}
\hdashline \specialrule{0em}{0pt}{0.02cm}
%\specialrule{0em}{0pt}{0.2pt}
WFG2 & 5 & 6.49$e-$02(4) & 6.73$e-$02(5)$-$ & 6.32$e-$02(3)+ & \textbf{5.76$e-$02(1)+} & 5.86$e-$02(2)+ & 9.22$e-$02(6)$-$ \\
WFG2 & 10 & 9.17$e-$02(4) & 9.29$e-$02(5)$-$ & 8.74$e-$02(2)+ & \textbf{8.15$e-$02(1)+} & 8.80$e-$02(3)= & 1.32$e-$01(6)$-$ \\
WFG2 & 15 & 1.12$e-$01(5) & 1.10$e-$01(4)= & 9.89$e-$02(3)+ & \textbf{8.65$e-$02(1)+} & 8.81$e-$02(2)+ & 1.21$e-$01(6)$-$ \\
%\specialrule{0em}{0pt}{0.2pt}
\hdashline \specialrule{0em}{0pt}{0.02cm}
%\specialrule{0em}{0pt}{0.2pt}
WFG3 & 5 & 8.65$e-$02(2) & 2.34$e-$01(5)$-$ & \textbf{5.12$e-$02(1)+} & 1.36$e-$01(3)$-$ & 1.57$e-$01(4)$-$ & 5.27$e-$01(6)$-$ \\
WFG3 & 10 & 4.32$e+$00(4) & 1.34$e+$01(5)$-$ & \textbf{1.28$e+$00(1)+} & 2.49$e+$00(2)+ & 3.06$e+$00(3)+ & 1.41$e+$02(6)$-$ \\
WFG3 & 15 & 1.30$e+$02(4) & 6.17$e+$02(5)$-$ & \textbf{3.64$e+$01(1)+} & 6.51$e+$01(2)+ & 8.16$e+$01(3)+ & 4.45$e+$03(6)$-$ \\
%\specialrule{0em}{0pt}{0.2pt}
\hdashline \specialrule{0em}{0pt}{0.02cm}
%\specialrule{0em}{0pt}{0.2pt}
WFG4 & 5 & 1.69$e-$01(3) & 1.69$e-$01(4)= & 1.73$e-$01(5)$-$ & 1.57$e-$01(2)+ & \textbf{1.51$e-$01(1)+} & 2.57$e-$01(6)$-$ \\
WFG4 & 10 & 3.65$e-$01(3) & 3.61$e-$01(2)+ & 3.74$e-$01(4)$-$ & 3.84$e-$01(5)$-$ & \textbf{3.47$e-$01(1)+} & 8.20$e-$01(6)$-$ \\
WFG4 & 15 & 5.38$e-$01(3) & 5.36$e-$01(2)+ & 5.57$e-$01(4)$-$ & \textbf{5.17$e-$01(1)+} & 5.85$e-$01(5)$-$ & 9.79$e-$01(6)$-$ \\
%\specialrule{0em}{0pt}{0.2pt}
\hdashline \specialrule{0em}{0pt}{0.02cm}
%\specialrule{0em}{0pt}{0.2pt}
WFG5 & 5 & 1.68$e-$01(4) & 1.65$e-$01(3)+ & 1.72$e-$01(5)$-$ & 1.57$e-$01(2)+ & \textbf{1.49$e-$01(1)+} & 2.42$e-$01(6)$-$ \\
WFG5 & 10 & 3.63$e-$01(3) & 3.53$e-$01(2)+ & 3.71$e-$01(4)$-$ & 3.79$e-$01(5)$-$ & \textbf{3.44$e-$01(1)+} & 7.48$e-$01(6)$-$ \\
WFG5 & 15 & 5.33$e-$01(3) & 5.26$e-$01(2)+ & 5.50$e-$01(4)$-$ & \textbf{5.12$e-$01(1)+} & 5.74$e-$01(5)$-$ & 9.32$e-$01(6)$-$ \\
%\specialrule{0em}{0pt}{0.2pt}
\hdashline \specialrule{0em}{0pt}{0.02cm}
%\specialrule{0em}{0pt}{0.2pt}
WFG6 & 5 & 1.72$e-$01(4) & 1.71$e-$01(3)= & 1.74$e-$01(5)$-$ & 1.63$e-$01(2)+ & \textbf{1.49$e-$01(1)+} & 2.93$e-$01(6)$-$ \\
WFG6 & 10 & 3.72$e-$01(3) & 3.59$e-$01(2)+ & 3.78$e-$01(4)$-$ & 3.93$e-$01(5)$-$ & \textbf{3.50$e-$01(1)+} & 8.99$e-$01(6)$-$ \\
WFG6 & 15 & 5.66$e-$01(4) & 5.34$e-$01(2)+ & 5.65$e-$01(3)= & \textbf{5.20$e-$01(1)+} & 5.77$e-$01(5)$-$ & 1.08$e+$00(6)$-$ \\
%\specialrule{0em}{0pt}{0.2pt}
\hdashline \specialrule{0em}{0pt}{0.02cm}
%\specialrule{0em}{0pt}{0.2pt}
WFG7 & 5 & 1.73$e-$01(3) & 1.73$e-$01(4)= & 1.75$e-$01(5)$-$ & 1.55$e-$01(2)+ & \textbf{1.52$e-$01(1)+} & 2.97$e-$01(6)$-$ \\
WFG7 & 10 & 3.72$e-$01(2) & \textbf{3.60$e-$01(1)+} & 3.81$e-$01(5)$-$ & 3.79$e-$01(4)$-$ & 3.74$e-$01(3)$-$ & 8.65$e-$01(6)$-$ \\
WFG7 & 15 & 5.57$e-$01(3) & 5.34$e-$01(2)+ & 5.63$e-$01(4)$-$ & \textbf{5.15$e-$01(1)+} & 5.97$e-$01(5)$-$ & 1.07$e+$00(6)$-$ \\
%\specialrule{0em}{0pt}{0.2pt}
\hdashline \specialrule{0em}{0pt}{0.02cm}
%\specialrule{0em}{0pt}{0.2pt}
WFG8 & 5 & 1.92$e-$01(3) & 1.91$e-$01(2)= & 1.93$e-$01(4)= & 1.97$e-$01(5)$-$ & \textbf{1.81$e-$01(1)+} & 2.75$e-$01(6)$-$ \\
WFG8 & 10 & \textbf{3.73$e-$01(1)} & 3.74$e-$01(2)= & 3.80$e-$01(3)$-$ & 4.65$e-$01(5)$-$ & 4.50$e-$01(4)$-$ & 9.13$e-$01(6)$-$ \\
WFG8 & 15 & 5.79$e-$01(3) & \textbf{5.54$e-$01(1)+} & 5.66$e-$01(2)+ & 5.92$e-$01(5)$-$ & 5.80$e-$01(4)= & 1.03$e+$00(6)$-$ \\
%\specialrule{0em}{0pt}{0.2pt}
\hdashline \specialrule{0em}{0pt}{0.02cm}
%\specialrule{0em}{0pt}{0.2pt}
WFG9 & 5 & 1.58$e-$01(3) & 1.59$e-$01(4)$-$ & 1.63$e-$01(5)$-$ & 1.53$e-$01(2)+ & \textbf{1.46$e-$01(1)+} & 2.58$e-$01(6)$-$ \\
WFG9 & 10 & 3.47$e-$01(3) & 3.47$e-$01(2)= & \textbf{3.42$e-$01(1)+} & 3.92$e-$01(5)$-$ & 3.57$e-$01(4)= & 8.02$e-$01(6)$-$ \\
WFG9 & 15 & 5.03$e-$01(2) & 5.14$e-$01(3)$-$ & \textbf{4.98$e-$01(1)+} & 5.38$e-$01(4)$-$ & 5.46$e-$01(5)$-$ & 8.73$e-$01(6)$-$ \\
\hline
    \multicolumn{2}{c}{Win/Tie/Loss} & - & 10/8/9 & 11/2/14 & 17/0/10 & 14/4/9 & 0/0/27 \\
\bottomrule
\end{tabular*}
\end{table*}

From the comparison results we find that, as analyzed above, the normalized SRA3 algorithm's solution set tends to extreme solutions on the DTLZ1 and DTLZ3 problems and thus performs poorly, but it performs well on DTLZ2 and DTLZ4 problems. Overall, on the DTLZ problems, SRA is the best-performing algorithm, and SRA3 has medium performance. And, while SRA3 does not perform well at the 5 objectives, it does significant improvement at the 10 and 15 objectives. In contrast, IBEA, which only uses the $I_{\epsilon+}$ indicator, has poor diversity in each problem, Two$\_$Arch2 performs very poorly in HV indicator on the DTLZ2 and DTLZ4 problems, and even has difficulty converging on the DTLZ3 problem, which requires a strong convergence capability of the algorithm, and NSGA-III encountered the same difficulty on the DTLZ3 problem, while MOEA/D has poor diversity on DTLZ4 problem and is not competitive. While on the WFG problems, normalization eliminates the effect of different objectives' inconsistent contributions to the indicators due to the inconsistent range of each objective, resulting in a significant improvement in the performance of SRA3. Overall, IBEA and NSGA-III are the two best performing algorithms on the WFG problems, while SRA3 is better than Two$\_$Arch2 and much better than SRA and MOEA/D. In general, SRA outperforms SRA3 on the DTLZ problems, but is worse than SRA3 on the WFG problems. IBEA and NSGA-III outperform SRA3 on the WFG problems, but IBEA performs poorly on the DTLZ problems, while NSGA-III may have trouble converging on DTLZ3 problem. SRA3, on the other hand, has medium performance but is more stable, has no obvious shortcomings, and has good convergence and diversity on all problems.

\subsection{Scalability analysis}

We found that SRA3 performs better on 10 and 15 objectives than on 5 objectives. Meanwhile, SRA uses the same two indicators, $I_{\epsilon+}$ and $I_{SDE}$, and Two$\_$Arch2 also uses a two-archive framework and its CA archive is also updated based on the $I_{\epsilon+}$ indicator. Not only that, in the original paper of Two$\_$Arch2, the authors compared the Two$\_$Arch2 algorithm with NSGA-III on the DTLZ problems with 20 objectives and proved that the Two$\_$Arch2 algorithm has better performance on 20 and 25 objectives \cite{wang2014two_arch2}. Therefore, this paper compared SRA3 with SRA and Two$\_$Arch2 on the DTLZ and WFG problems with 20 objectives. To be fair, all algorithms were normalized and the results are shown in Tables 16 to 19.

From the comparison results, it can be seen that for the problems with 20 and 25 objectives, SRA3 is significantly better than SRA and Two$\_$Arch2 on the DTLZ problems, and for the WFG problems, SRA3 is better than SRA and performs comparably to Two$\_$Arch2, with Two$\_$Arch2 performing better on the IGD indicator, but SRA3 performing better on the HV indicator. Meanwhile, in order to explore the performance trends of several algorithms as the number of objectives increases, we analyzed the trends of HV and IGD indicators on several algorithms as the number of objectives increases on DTLZ2, DTLZ3, WFG7, and WFG9 problems, as shown in Figure 9 (the IGD indicator of Two$\_$Arch2 algorithm on the DTLZ3 problem is not shown here, because it has difficulty converging on the DTLZ3 problem with more than 15 objectives, so the IGD value is very poor). 

Figure 9 shows that when the number of objectives increases, SRA3 gradually approaches or widens the gap with the other algorithms in terms of HV and IGD indicators. When the number of objectives reaches 20 and 25, SRA3 becomes the best-performing algorithm on most problems, especially in the HV indicator, which is significantly better than other algorithms. Therefore, we believe that SRA3 is more competitive than other algorithms as the number of objectives increases.

\section{Conclusion}

In this paper, a two-archive based multi-indicator multi-objective optimization algorithm (SRA3) is proposed, which can perform adaptive parental selection based on the ratio of non-dominated solutions in CA and DA archives without setting additional parameters. And in the environment selection, compared with other multi-indicator multi-objective optimization algorithms, SRA3 is equivalent to two single-indicator-based algorithms for next-generation parental selection, without considering multiple indicators inconsistent with each other, and thus has obvious efficiency advantages. Meanwhile, SRA3 was normalized to ensure that each objective have the same contribution to the indicators. Then the performance of SRA3 before and after normalization is compared and the final results indicate that the normalized SRA3 is significantly better than the one before normalization in the majority of problems. Subsequently, this paper explores the effect of normalization on the IB-MOEAs and finds that although normalization makes the $I_{\epsilon+}$ indicators prefer extreme solutions and makes the solution set tend to be marginalized, it eliminates the effect of the different contribution of each objective to the indicators due to the different objective ranges and makes it easier to find extreme solutions. Therefore, normalization should be performed when the objective ranges of the problem are inconsistent. But if the range of each objective is consistent, the algorithm should be normalized if it is difficult to find extreme solutions for the problem, otherwise, normalization is not necessary. Meanwhile, in this paper, we compare SRA3 with several state-of-the-art algorithms on DTLZ and WFG problems with 5, 10, and 15 objectives and find that SRA3 has better convergence and diversity, and performs consistently. Then, to explore the performance of SRA3 in higher dimensions and how the algorithm's performance changes as the number of objectives increases, we compare the normalized present algorithm with SRA and Two$\_$Arch2 on DTLZ and WFG problems with 20 and 25 objectives and find that SRA3 has advantages in the case of a larger number of objectives. Moreover, as the number of objective dimensions increases SRA3 can gradually approach or widen the gap with other algorithms; that is, as the number of objectives increases, this algorithm becomes more competitive.

Although the overall performance of SRA3 is satisfactory, more research is needed in the future. Firstly, the $I_{\epsilon+}$ indicator prefers extreme solutions after normalization, so can we improve the $I_{\epsilon+}$ indicator to alleviate this preference or use other combinations of indicators to get better results. Second, compared to the CA archive, the DA archive seems to be underutilized, so is there a better approach to combine the CA and DA archives? Finally, the performance of this algorithm still needs to be verified in more real-world problems.

\begin{table}[width=.9\linewidth,cols=5,pos=H]
\caption{Performance comparison of SRA3, SRA, and Two$\_$Arch2 in terms of the average HV values on DTLZ problems with 20 and 25 objectives. The indicator values are followed by their comparison result with the Wilcoxon rank sum test of SRA3 (significance level = 0.05). And the best result is highlighted in boldface.}\label{tbl1}
\begin{tabular*}{\tblwidth}{@{} CCCCC@{} }
\toprule
Problem & $m$ & SRA3$_{norm}$ & SRA$_{norm}$ & Two$\_$Arch2$_{norm}$ \\
\midrule
DTLZ1 & 20 & 8.61$e-$01 &  8.28$e-$01= & \textbf{9.45$e-$01}+ \\
DTLZ1 & 25 & 8.50$e-$01 &  8.03$e-$01$-$ & \textbf{8.88$e-$01}+ \\
%\specialrule{0em}{0pt}{0.2pt}
\hdashline \specialrule{0em}{0pt}{0.02cm}
%\specialrule{0em}{0pt}{0.2pt}
DTLZ2 & 20 & \textbf{9.94$e-$01} &  9.28$e-$01$-$ & 4.36$e-$01$-$ \\
DTLZ2 & 25 & \textbf{9.96$e-$01} &  9.24$e-$01$-$ & 3.09$e-$01$-$ \\
%\specialrule{0em}{0pt}{0.2pt}
\hdashline \specialrule{0em}{0pt}{0.02cm}
%\specialrule{0em}{0pt}{0.2pt}
DTLZ3 & 20 & \textbf{7.24$e-$01} &  4.27$e-$01$-$ & 0.00$e-$00$-$ \\
DTLZ3 & 25 & \textbf{7.02$e-$01} &  4.37$e-$01$-$ & 0.00$e-$00$-$ \\
%\specialrule{0em}{0pt}{0.2pt}
\hdashline \specialrule{0em}{0pt}{0.02cm}
%\specialrule{0em}{0pt}{0.2pt}
DTLZ4 & 20 & \textbf{9.96$e-$01} &  9.76$e-$01$-$ & 5.24$e-$01$-$ \\
DTLZ4 & 25 & \textbf{9.97$e-$01} &  9.78$e-$01$-$ & 4.61$e-$01$-$ \\
\hline
    \multicolumn{2}{c}{Win/Tie/Loss} & - & 0/1/7 & 2/0/6\\
\bottomrule
\end{tabular*}
\end{table}

\begin{table}[width=.9\linewidth,cols=5,pos=H]
\caption{Performance comparison of SRA3, SRA, and Two$\_$Arch2 in terms of the average IGD values on DTLZ problems with 20 and 25 objectives. The indicator values are followed by their comparison result with the Wilcoxon rank sum test of SRA3 (significance level = 0.05). And the best result is highlighted in boldface.}\label{tbl1}
\begin{tabular*}{\tblwidth}{@{} CCCCC@{} }
\toprule
Problem & $m$ & SRA3$_{norm}$ & SRA$_{norm}$ & Two$\_$Arch2$_{norm}$ \\
\midrule
DTLZ1 & 20 & 5.81$e-$01 &  5.75$e-$01= & \textbf{2.91$e-$01}+ \\
DTLZ1 & 25 & 5.97$e-$01 &  6.23$e-$01= & \textbf{3.37$e-$01}+ \\
%\specialrule{0em}{0pt}{0.2pt}
\hdashline \specialrule{0em}{0pt}{0.02cm}
%\specialrule{0em}{0pt}{0.2pt}
DTLZ2 & 20 & \textbf{6.60$e-$01} &  6.73$e-$01$-$ & 6.77$e-$01$-$ \\
DTLZ2 & 25 & \textbf{6.86$e-$01} &  7.27$e-$01$-$ & 7.86$e-$01$-$ \\
%\specialrule{0em}{0pt}{0.2pt}
\hdashline \specialrule{0em}{0pt}{0.02cm}
%\specialrule{0em}{0pt}{0.2pt}
DTLZ3 & 20 & \textbf{8.39$e-$01} &  9.76$e-$01$-$ & 1.27$e$+01$-$ \\
DTLZ3 & 25 & \textbf{8.95$e-$01} &  1.02$e+$00$-$ & 1.20$e$+01$-$ \\
%\specialrule{0em}{0pt}{0.2pt}
\hdashline \specialrule{0em}{0pt}{0.02cm}
%\specialrule{0em}{0pt}{0.2pt}
DTLZ4 & 20 & 6.61$e-$01 &  6.42$e-$01+ & \textbf{6.30$e-$01}+ \\
DTLZ4 & 25 & \textbf{6.85$e-$01} &  6.93$e-$01$-$ & 7.08$e-$01$-$ \\
\hline
    \multicolumn{2}{c}{Win/Tie/Loss} & - & 1/2/5 & 3/0/5\\
\bottomrule
\end{tabular*}
\end{table}

\begin{table}[width=.9\linewidth,cols=5,pos=H]
\caption{Performance comparison of SRA3, SRA, and Two$\_$Arch2 in terms of the average HV values on WFG problems with 20 and 25 objectives. The indicator values are followed by their comparison result with the Wilcoxon rank sum test of SRA3 (significance level = 0.05). The best result is highlighted in boldface.}\label{tbl1}
\begin{tabular*}{\tblwidth}{@{} CCCCC@{} }
\toprule
Problem & $m$ & SRA3$_{norm}$ & SRA$_{norm}$ & Two$\_$Arch2$_{norm}$ \\
\midrule
WFG1 & 20 & 9.93$e-$01 &  9.93$e-$01$=$ & \textbf{9.98$e-$01}+ \\
WFG1 & 25 & 9.93$e-$01 &  9.93$e-$01$=$ & \textbf{9.98$e-$01}+ \\
%\specialrule{0em}{0pt}{0.2pt}
\hdashline \specialrule{0em}{0pt}{0.02cm}
%\specialrule{0em}{0pt}{0.2pt}
WFG2 & 20 & 9.92$e-$01 &  9.89$e-$01$-$ & \textbf{9.97$e-$01}+ \\
WFG2 & 25 & 9.90$e-$01 &  9.88$e-$01$-$ & \textbf{9.96$e-$01}+ \\
%\specialrule{0em}{0pt}{0.2pt}
\hdashline \specialrule{0em}{0pt}{0.02cm}
%\specialrule{0em}{0pt}{0.2pt}
WFG3 & 20 & 0.00$e-$00 &  0.00$e-$00= & 0.00$e-$00= \\
WFG3 & 25 & 0.00$e-$00 &  0.00$e-$00= & 0.00$e-$00= \\
%\specialrule{0em}{0pt}{0.2pt}
\hdashline \specialrule{0em}{0pt}{0.02cm}
%\specialrule{0em}{0pt}{0.2pt}
WFG4 & 20 & \textbf{8.90$e-$01} &  8.37$e-$01$-$ & 7.74$e-$01$-$ \\
WFG4 & 25 & \textbf{9.05$e-$01} &  8.81$e-$01$=$ & 7.64$e-$01$-$ \\
%\specialrule{0em}{0pt}{0.2pt}
\hdashline \specialrule{0em}{0pt}{0.02cm}
%\specialrule{0em}{0pt}{0.2pt}
WFG5 & 20 & \textbf{7.92$e-$01} &  7.32$e-$01$-$ & 6.20$e-$01$-$ \\
WFG5 & 25 & \textbf{7.41$e-$01} &  7.14$e-$01$-$ & 5.91$e-$01$-$ \\
%\specialrule{0em}{0pt}{0.2pt}
\hdashline \specialrule{0em}{0pt}{0.02cm}
%\specialrule{0em}{0pt}{0.2pt}
WFG6 & 20 & \textbf{8.48$e-$01} &  7.31$e-$01$-$ & 6.79$e-$01$-$ \\
WFG6 & 25 & \textbf{8.36$e-$01} &  7.06$e-$01$-$ & 6.58$e-$01$-$ \\
%\specialrule{0em}{0pt}{0.2pt}
\hdashline \specialrule{0em}{0pt}{0.02cm}
%\specialrule{0em}{0pt}{0.2pt}
WFG7 & 20 & \textbf{9.28$e-$01} &  8.61$e-$01$-$ & 7.45$e-$01$-$ \\
WFG7 & 25 & \textbf{9.15$e-$01} &  8.72$e-$01$-$ & 7.17$e-$01$-$ \\
%\specialrule{0em}{0pt}{0.2pt}
\hdashline \specialrule{0em}{0pt}{0.02cm}
%\specialrule{0em}{0pt}{0.2pt}
WFG8 & 20 & \textbf{9.07$e-$01} &  7.36$e-$01$-$ & 5.17$e-$01$-$ \\
WFG8 & 25 & \textbf{9.53$e-$01} &  8.25$e-$01$-$ & 4.87$e-$01$-$ \\
%\specialrule{0em}{0pt}{0.2pt}
\hdashline \specialrule{0em}{0pt}{0.02cm}
%\specialrule{0em}{0pt}{0.2pt}
WFG9 & 20 & \textbf{7.43$e-$01} &  7.40$e-$01$=$ & 5.83$e-$01$-$ \\
WFG9 & 25 & \textbf{7.30$e-$01} &  7.13$e-$01$=$ & 5.89$e-$01$-$ \\
\hline
    \multicolumn{2}{c}{Win/Tie/Loss} & - & 0/7/11 & 4/2/12\\
\bottomrule
\end{tabular*}
\end{table}

\begin{table}[width=.9\linewidth,cols=5,pos=H]
\caption{Performance comparison of SRA3, SRA, and Two$\_$Arch2 in terms of the average IGD values on WFG problems with 20 and 25 objectives. The indicator values are followed by their comparison result with the Wilcoxon rank sum test of SRA3 (significance level = 0.05). The best result is highlighted in boldface.}\label{tbl1}
\begin{tabular*}{\tblwidth}{@{} CCCCC@{} }
\toprule
Problem & $m$ & SRA3$_{norm}$ & SRA$_{norm}$ & Two$\_$Arch2$_{norm}$ \\
\midrule
WFG1 & 20 & 1.68$e-$01 &  1.64$e-$01$+$ & \textbf{1.52$e-$01}+ \\
WFG1 & 25 & 1.33$e-$01 &  1.29$e-$01$+$ & \textbf{1.21$e-$01}+ \\
%\specialrule{0em}{0pt}{0.2pt}
\hdashline \specialrule{0em}{0pt}{0.02cm}
%\specialrule{0em}{0pt}{0.2pt}
WFG2 & 20 & 1.79$e-$01 &  1.80$e-$01$=$ & \textbf{1.61$e-$01$+$} \\
WFG2 & 25 & 1.36$e-$01 &  1.34$e-$01$=$ & \textbf{1.25$e-$01$+$} \\
%\specialrule{0em}{0pt}{0.2pt}
\hdashline \specialrule{0em}{0pt}{0.02cm}
%\specialrule{0em}{0pt}{0.2pt}
WFG3 & 20 & 3.51$e$+03 & 1.81$e$+04$-$ & \textbf{2.42$e$+03}= \\
WFG3 & 25 & 1.16$e$+05 &5.35$e$+05$-$ & \textbf{7.04$e$+03}+ \\
%\specialrule{0em}{0pt}{0.2pt}
\hdashline \specialrule{0em}{0pt}{0.02cm}
%\specialrule{0em}{0pt}{0.2pt}
WFG4 & 20 & 6.26$e-$01 &  6.13$e-$01$+$ & \textbf{5.92$e-$01}+ \\
WFG4 & 25 & 6.73$e-$01 &  6.75$e-$01$=$ & \textbf{6.49$e-$01}+ \\
%\specialrule{0em}{0pt}{0.2pt}
\hdashline \specialrule{0em}{0pt}{0.02cm}
%\specialrule{0em}{0pt}{0.2pt}
WFG5 & 20 & 6.05$e-$01 &  5.99$e-$01$+$ & \textbf{5.88$e-$01}+ \\
WFG5 & 25 & \textbf{6.36$e-$01} &  6.47$e-$01$-$ & 6.46$e-$01$-$ \\
%\specialrule{0em}{0pt}{0.2pt}
\hdashline \specialrule{0em}{0pt}{0.02cm}
%\specialrule{0em}{0pt}{0.2pt}
WFG6 & 20 & 6.57$e-$01 &  6.11$e-$01$+$ & \textbf{5.97$e-$01}+ \\
WFG6 & 25 & 7.02$e-$01 &  6.64$e-$01$+$ & \textbf{6.55$e-$01}+ \\
%\specialrule{0em}{0pt}{0.2pt}
\hdashline \specialrule{0em}{0pt}{0.02cm}
%\specialrule{0em}{0pt}{0.2pt}
WFG7 & 20 & 6.36$e-$01 &  6.10$e-$01$+$ & \textbf{5.94$e-$01}+ \\
WFG7 & 25 & 6.65$e-$01 &  6.76$e-$01$-$ & \textbf{6.60$e-$01}= \\
%\specialrule{0em}{0pt}{0.2pt}
\hdashline \specialrule{0em}{0pt}{0.02cm}
%\specialrule{0em}{0pt}{0.2pt}
WFG8 & 20 & 6.78$e-$01 &  \textbf{6.59$e-$01$+$} & 6.71$e-$01= \\
WFG8 & 25 & \textbf{7.35$e-$01} &  7.38$e-$01$=$ & 7.41$e-$01= \\
%\specialrule{0em}{0pt}{0.2pt}
\hdashline \specialrule{0em}{0pt}{0.02cm}
%\specialrule{0em}{0pt}{0.2pt}
WFG9 & 20 & \textbf{5.75$e-$01} &  5.91$e-$01$-$ & 6.13$e-$01$-$ \\
WFG9 & 25 & \textbf{6.38$e-$01} &  6.50$e-$01$-$ & 6.76$e-$01$-$ \\
\hline
    \multicolumn{2}{c}{Win/Tie/Loss} & - & 8/4/6 & 11/4/3 \\
\bottomrule
\end{tabular*}
\end{table}

\begin{figure*}[H]
    \centering
    \begin{subfigure}{4cm}
    \includegraphics[width=4cm]{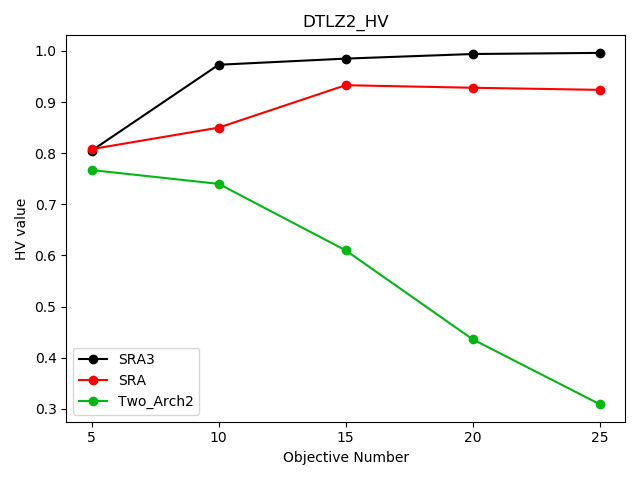}
    \subcaption*{DTLZ2-HV}
    \end{subfigure}
    \begin{subfigure}{4cm}
    \includegraphics[width=4cm]{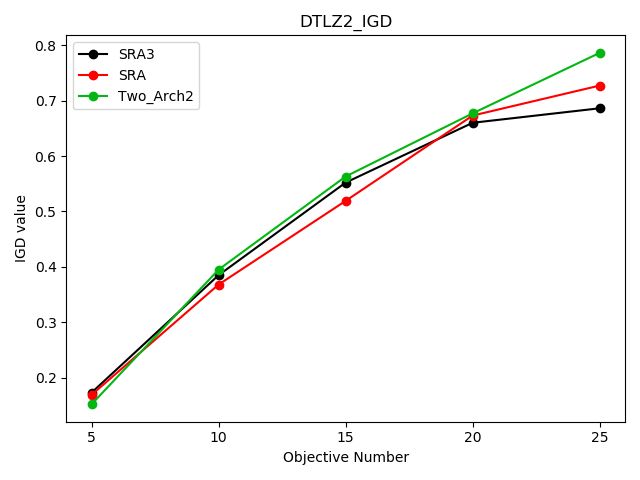}
    \subcaption*{DTLZ2-IGD}
    \end{subfigure}
    \begin{subfigure}{4cm}
    \includegraphics[width=4cm]{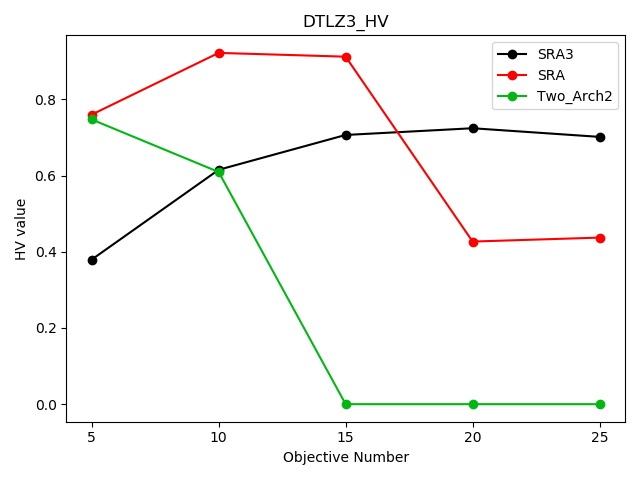}
    \subcaption*{DTLZ3-HV}
    \end{subfigure}
    \begin{subfigure}{4cm}
    \includegraphics[width=4cm]{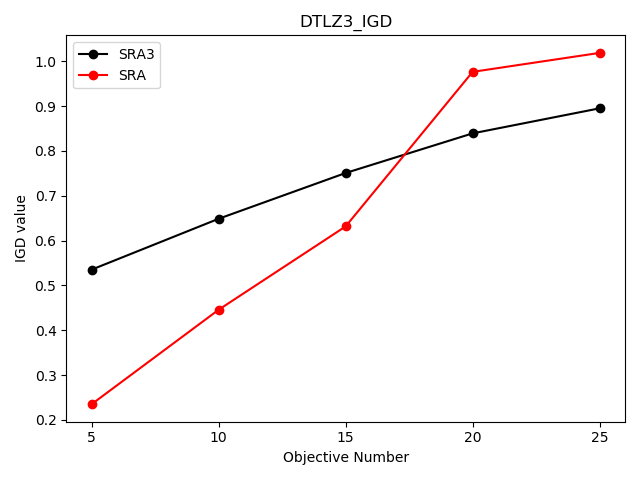}
    \subcaption*{DTLZ3-IGD}
    \end{subfigure}
    \begin{subfigure}{4cm}
    \includegraphics[width=4cm]{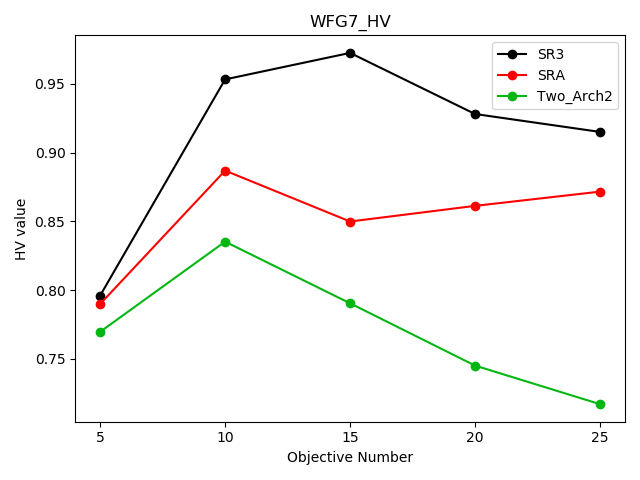}
    \subcaption*{WFG7-HV}
    \end{subfigure}
    \begin{subfigure}{4cm}
    \includegraphics[width=4cm]{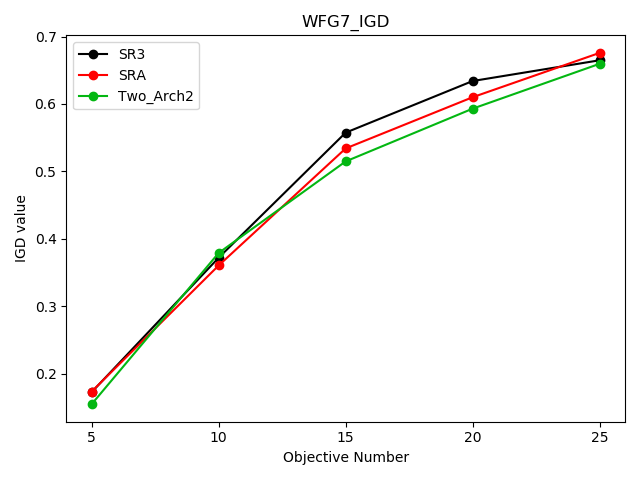}
    \subcaption*{WFG7-IGD}
    \end{subfigure}
    \begin{subfigure}{4cm}
    \includegraphics[width=4cm]{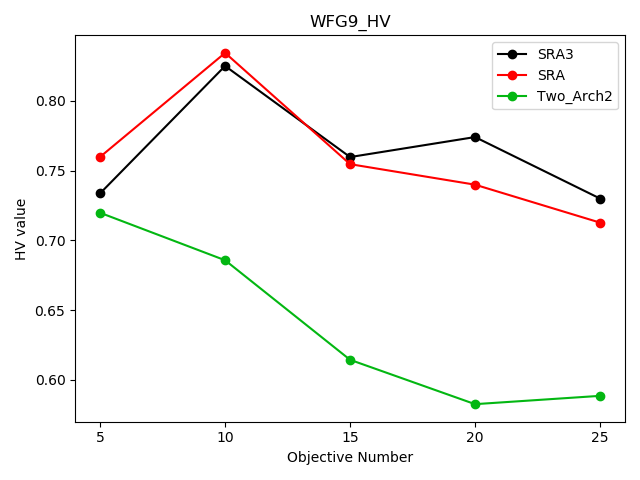}
    \subcaption*{WFG9-HV}
    \end{subfigure}
    \begin{subfigure}{4cm}
    \includegraphics[width=4cm]{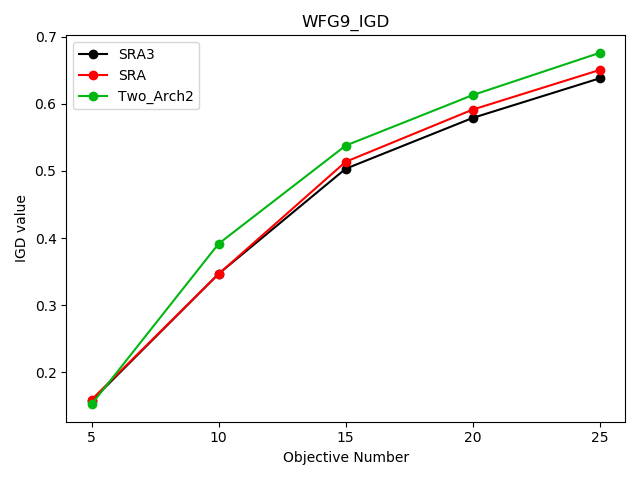}
    \subcaption*{WFG9-IGD}
    \end{subfigure}
    \caption{The trend of HV and IGD indicators of each algorithm on the DTLZ2, DTLZ3, WFG7, WFG9 problems with increasing number of objectives}
\end{figure*}

\printcredits

\bibliographystyle{model1-num-names}
% \bibliographystyle{cas-model2-names}

\bibliography{cas-refs}

\end{document}